\documentclass[10pt,twocolumn,letterpaper]{article}

\usepackage{cvpr}
\usepackage[accsupp]{axessibility}

\usepackage{bbm}
\usepackage{amsmath}
\usepackage{graphicx}
\usepackage{multirow}

\usepackage{amsmath,amsfonts,bm}

\def\eqref#1{equation~\ref{#1}}
\def\1{\bm{1}}

\def\rveta{{\bm{\eta}}}

\def\vw{{\bm{w}}}
\def\vx{{\bm{x}}}

\def\vz{{\bm{z}}}

\def\mI{{\bm{I}}}

\DeclareMathAlphabet{\mathsfit}{\encodingdefault}{\sfdefault}{m}{sl}
\SetMathAlphabet{\mathsfit}{bold}{\encodingdefault}{\sfdefault}{bx}{n}

\def\gN{{\mathcal{N}}}

\def\gX{{\mathcal{X}}}
\def\gY{{\mathcal{Y}}}

\def\sP{{\mathbb{P}}}

\def\sS{{\mathbb{S}}}

\newcommand{\E}{\mathbb{E}}

\DeclareMathOperator*{\argmax}{arg\,max}

\definecolor{cvprblue}{rgb}{0.21,0.49,0.74}
\usepackage[pagebackref,breaklinks,colorlinks,allcolors=cvprblue]{hyperref}

\def\paperID{8734} % *** Enter the Paper ID here
\def\confName{CVPR}
\def\confYear{2025}

\title{Uncertainty Weighted Gradients for Model Calibration}

\author{
 Jinxu~Lin$^{1}$ \ \ \ \ Linwei~Tao$^{1}$ \ \ \ \ Minjing~Dong$^{2}$ \ \ \ \ Chang~Xu$^{1}$
 \and $^{1}$The University of Sydney, \ $^{2}$City University of Hong Kong
 \and {\tt\small \{jinxu.lin, linwei.tao, c.xu\}@sydney.edu.au, minjdong@cityu.edu.hk}
}

\begin{document}
\maketitle
\begin{abstract}
Model calibration is essential for ensuring that the predictions of deep neural networks accurately reflect true probabilities in real-world classification tasks. However, deep networks often produce over-confident or under-confident predictions, leading to miscalibration. Various methods have been proposed to address this issue by designing effective loss functions for calibration, such as focal loss. In this paper, we analyze its effectiveness and provide a unified loss framework of focal loss and its variants, where we mainly attribute their superiority in model calibration to the loss weighting factor that estimates sample-wise uncertainty. Based on our analysis, existing loss functions fail to achieve optimal calibration performance due to two main issues: including misalignment during optimization and insufficient precision in uncertainty estimation. Specifically, focal loss cannot align sample uncertainty with gradient scaling and the single logit cannot indicate the uncertainty. To address these issues, we reformulate the optimization from the perspective of gradients, which focuses on uncertain samples. Meanwhile, we propose using the Brier Score as the loss weight factor, which provides a more accurate uncertainty estimation via all the logits. Extensive experiments on various models and datasets demonstrate that our method achieves state-of-the-art (SOTA) performance.\footnote{Code is available at \href{https://github.com/Jinxu-Lin/BSCE-GRA}{https://github.com/Jinxu-Lin/BSCE-GRA}.}
\end{abstract}
\section{Introduction}
\label{sec:intro}
Deep Neural Networks (DNNs) have achieved remarkable success in various domains, including image classification.
However, recent studies~\cite{guo2017calibration,tao2024a} reveal that DNNs often suffer from mis-calibration in classification task, exhibiting over-confidence or under-confidence in their predictions.
For example, a model might output a confidence score of 0.8 for a particular prediction, which does not necessarily correspond to an 80\% probability of correctness. 

Calibrating DNNs, which aligns the predicted confidence with the true probabilities, is therefore crucial to enhancing their reliability in practical applications, such as autonomous driving, medical imaging, and weather forecasting.
To address mis-calibration issues, several methods~\cite{tao2023calibrating,tao2024feature} have been proposed, many of which focus on modifying the loss function during training.
A common approach involves adding regularization terms with the Cross-Entropy (CE) loss to improve calibration, as seen in methods like Maximum Mean Calibration Error (MMCE)~\cite{kumar2018trainable} and Meta Calibration~\cite{bohdal2021meta}.
Additionally, Focal Loss (FL)~\cite{lin2017focal}, which adjusts per-sample loss weights based on prediction difficulty, has been shown to improve calibration performance. \citet{mukhoti2020calibrating} attribute the effectiveness of FL to its implicit regularization on the entropy of predicted probabilities, which mitigates overconfidence. Similarly, Dual Focal Loss (DFL)~\cite{tao2023dual} incorporates the second-most-probable class in the weighting mechanism to address model under-confidence.
Another research interest regarding calibration is the evaluation of calibration performance.
Metrics commonly employed include Expected Calibration Error (ECE)\cite{guo2017calibration} and Brier Score\cite{brier1950verification}.

The focal-loss-based methods can be unified under a general loss framework expressed as \( u \cdot \text{CE} \), where \( u \) is a loss weighting factor.
In FL, the factor $u_{\text{\tiny FL}}$ can be treated as an estimation for sample-wise difficulty .
Interestingly, we observe that $u_{\text{\tiny FL}}$, originally designed for binary classification, is mathematically equivalent to the Brier Score for binary cases, which can measure the uncertainty.
Furthermore, the weighting term $u_{\text{\tiny DFL}}$ in DFL aligns with the Brier Score in three-class classification scenarios.
This observation motivates the hypothesis that weighting the loss with sample calibration metrics, such as the Brier Score, could better identify uncertain samples and facilitate targeted training.

Furthermore, our investigation reveals a limitation of directly applying the weighting factor \( u \) to the loss term.
Specifically, the vanilla optimization of focal loss would achieve a misalignment with the objective of scaling gradients for harder samples with larger magnitudes.
Based on our analysis, we mainly attribute this issue to a differentiable loss weighting factor \( u \), which could disrupt the positive correlation between the CE and its gradient magnitude during backpropagation, impeding the model's ability to prioritize higher-uncertain samples effectively.

To address these issues, we first propose applying the weighting factor to scaling the gradient rather than the loss itself.
This allows us directly aligns the gradient optimization with sample uncertainty, which ensures that harder samples receive appropriately scaled updates without disrupting the optimization process.
Within this framework, we then introduce a generalized form of the Brier Score as a gradient weighting factor, which provides a more accurate estimation of sample uncertainty that considers all categories.
This leads to the development of the BSCE-GRA loss function, which adjusts gradient to directly scale optimization based on sample uncertainty by Brier Score.

We conduct extensive experiments to validate the effectiveness of the proposed method, achieving state-of-the-art results across various datasets and model architectures, which demonstrate the effectiveness of our methods.
The contributions of this paper are summarized as:

\begin{enumerate}
\item We provide a new perspective by unifying some existing loss modification techniques for model calibration under a sample-weighting framework.
With this framework, we provide extensive analysis of their limitations.
\item We propose a simple yet effective optimization framework for model calibration with a gradient weighting factor, where we scale the gradients to encourage the model to focus on uncertain samples effectively.
\item We analyze the use of different uncertainty metrics within this framework and introduce BSCE-GRA, a new loss function based on the Brier Score that provides an accurate sample-wise uncertainty estimation.
\item We conduct extensive experiments under different settings to validate our proposed methods.
Our uncertainty-weighted framework shows consistent effectiveness across various uncertainty metrics, and the proposed BSCE-GRA loss achieves state-of-the-art results.
\end{enumerate}
\section{Related Works}
\label{sec: related}

In recent years, numerous techniques have been proposed to address the problem of network miscalibration, which can generally be classified into three categories.

The first category is post-hoc calibration techniques, which adjust model predictions after training by optimizing additional parameters on a held-out validation set.
These methods include Platt Scaling~\citep{platt1999probabilistic}, which performs a linear transformation on the original prediction logits; Isotonic Regression~\citep{zadrozny2002transforming}, which uses piecewise functions to transform logits; Bayesian Binning into Quantiles (BBQ)\citep{naeini2015obtaining}, which extends histogram binning with Bayesian model averaging; and Beta Calibration\citep{kull2017beta}, initially designed for binary classification and later generalized to multi-class settings with Dirichlet distributions by \citet{kull2019beyond}.
Temperature Scaling~\citep{guo2017calibration}, one of the most widely used post-hoc calibration methods, tunes the temperature parameter in the SoftMax function to minimize negative log-likelihood on a held-out validation set.
In this work, we report calibration performance with post-temperature scaling results.

The second category includes regularization techniques, which are known to effectively calibrate DNNs.
Data augmentation methods, such as Mixup~\citep{thulasidasan2019mixup} and AugMix~\citep{hendrycks*2020augmix}, train DNNs on mixed samples to mitigate overconfident predictions.
Model ensemble techniques, which involve independently training multiple DNNs and averaging their predictions, have been shown to enhance both accuracy and predictive uncertainty by aggregating outputs from multiple models~\citep{lakshminarayanan2017simple,zhang2020mix,rahaman2021uncertainty}.
Label smoothing~\citep{muller2019does}, which replaces one-hot labels with soft labels, encourages the model to make less confident predictions, thereby reducing overconfidence.
Additionally, weight decay has also been demonstrated to improve confidence calibration~\citep{guo2017calibration}.

The third category focuses on modifying the training loss to improve calibration.
These methods include adding a differentiable auxiliary surrogate loss for Expected Calibration Error (ECE)\citep{karandikar2021soft,krishnan2020improving,bohdal2021meta} or replacing the training loss with other loss functions, such as Mean Squared Error (MSE)\citep{hui2021evaluation}, Focal Loss~\citep{mukhoti2020calibrating}, Inverse Focal Loss~\citep{wang2021rethinking}, and Dual Focal Loss~\citep{tao2023dual}.
Among these, Focal Loss~\citep{mukhoti2020calibrating}, which adds a modulation term to the Cross-Entropy loss to focus on hard-to-classify samples, provides a simple and effective way to train calibrated models.
Focal Loss and Dual Focal Loss can be categorized as margin-based losses, similar to Hinge Loss~\citep{cortes1995support}, Triplet Loss~\citep{schroff2015facenet}, and Margin Ranking Loss~\citep{weston2011wsabie}.
\section{Preliminary}
\label{sec: preliminary}

\subsection{Problem Formulation}
\label{subsec: Calibration Error and Uncertainty}
Given a $K$-class dataset $\sS=\{\vz^{(1)},...,\vz^{(N)}\}$, where each training sample $\vz^{(n)}:=(x^{(n)},y^{(n)})$ is an input-label pair.
Let $\gX$ and $\gY$ represent the input space and the label space, respectively.
The ground truth label $y\in\gY$ is encoded in a one-hot vector format, where $y_i=1$ if $i\in K$ represents the actual class.
We define a classifier $f_\theta$ trained on $\sS$ that maps an input $\vx\in\gX$ to a probability distribution $\hat{p}(\vx)$.
The classifier then provides a prediction $k=\argmax_{i\in K}\hat{p}_{i}$, which indicates the index of the predicted label.
The predicted label $\hat{y}\in\gY$ is similarly represented as $y$ in the one-hot encoding of prediction $k$.
The confidence $\hat{p}_c$ is defined as the predicted probability associated with the prediction $k$.

A well-calibrated model ensures that the provided confidence $\hat{p}_c$ accurately reflects the true probability of correct classification.
We define the true class-posterior probability vector as $\rveta(\vx)=[\eta_1(\vx),...,\eta_K(\vx)]$, where $\eta_k(\vx)=\sP(y=k\vert \vx)$ is the true probability of class $k$ given input $\vx$.
Formally, a network is considered perfectly calibrated if $\sP(\hat{y}=y\vert\hat{p}_c=p)=p$ for all $p\in[0,1]$~\citep{guo2017calibration}, which can be also written as $\hat{p}(\vx)=\rveta(\vx)$.
The uncertainty of a model for a sample $\vx$, also referred to the calibrated error $c(\vx)$, can be computed as the difference between $\rveta(\vx)$ and $\hat{p}(\vx)$:
\begin{equation}
    c(\vx) = \Vert \eta(\vx) - \hat{p}_c(\vx) \Vert.
    \label{eq: calibrated error}
\end{equation}

\subsection{Metric for Calibrated Error}
In practice, it is hard to access the ground truth probability $\rveta(\vx)$ on real-world samples and thus the calibration error cannot be directly computed by Eq.~\ref{eq: calibrated error}, some alternative methods have been proposed to evaluate the uncertainty.

\noindent\textbf{ECE.}
The Expected Calibration Error (ECE) is defined as $\E_{\hat{p}_c}[\vert\sP(\hat{y}=y\vert\hat{p}_c)-\hat{p}_c\vert]$.
\citet{guo2017calibration} propose an approximation of ECE.
Specifically, all samples are divided into $M$ bins $\{B_m\}^M_{m=1}$ of equal width based on their confidence, where each bin $B_m$ contains all samples whose confidences fall within the range $\hat{p}_c\in[\frac{m}{M},\frac{m+1}{M})$.
For each bin $B_m$, the average confidence is computed as $C_m=\frac{1}{\vert B_m \vert}\sum_{i\in B_m}\hat{p}_c^{(i)}$ and the bin accuracy is computed as $A_m=\frac{1}{\vert B_m \vert}\sum_{i\in B_m}\mathbbm{1}(\hat{y}_k^{(i)}=y_k^{(i)})$, where $\mathbbm{1}$ is the indicator function.
The ECE can then be computed as the average $L_1$ difference between bin accuracy and confidence:
\begin{equation}
    \text{ECE} = \sum^M_{m=1}\frac{\vert B_m\vert}{N}\vert A_m - C_m \vert,
    \label{eq: definition of ece}
\end{equation}
where $N$ denotes the number of samples in each bin.
In addition to the ECE in Eq.\ref{eq: definition of ece}, there are several variants used to measure calibration error.
For example, AdaECE~\citep{nguyen2015posterior} groups samples into bins $B_m$ with an equal number of smaples such that $\vert B_m \vert = \vert B_n \vert$ for all bins.
Another variant, ClasswiesECE~\citep{kull2019beyond} measure the calibration error by considering each of the $K$ classes separately.

\noindent\textbf{Brier Score.}
Besides ECE computing the expectation of calibrated error among whole datasets, some other techniques have also been used to evaluate the sample-wise uncertainty.
Accuracy and calibration are distinct concepts, and one cannot be inferred from the other unequivocally.
The Brier Score (BS)~\citep{brier1950verification} unify these two concepts, which is commonly used in calibration literature.
It has been shown that this family of metrics can be decomposed into a calibration term and a refinement term.
Achieving an optimal score requires both accurate predictions and appropriate confidence levels.
For a given sample, the Brier Score is mathematically defined as the Mean Squared Error between the predicted probability distribution $\hat{p}$ and one-hot encoded ground truth label $y$.
We introduce a generalized Brier Score (gBS) form as follows:
\begin{equation}
    \text{gBS} = \sum_{i=1}^K\Vert\hat{p}_i(\vx)-y_i\Vert^\gamma_\beta
    \label{eq: definition of general brier score}
\end{equation}
where $\gamma$ and $\beta$ are the hyperparamters for the exponent and norm order, respectively.
When $\gamma=2$ and $\beta=2$, this formulation reduces to the original Brier Score.

\subsection{Calibrating Method}
\label{subsec: Calibrating Method}

\textbf{Temperature Scaling}
A widely used post-hoc technique for improving classification calibration is temperature scaling. It adjusts the sharpness of the output probability distribution by scaling the logits in the SoftMax function using a temperature parameter, defined as $\hat{p}_i=\frac{\exp(\hat{g}_i/T)}{\sum_{k=1}^K\exp(\hat{g}_k/T)}$, where $\hat{g}$ represents the logits before applying the SoftMax function, and $T$ is the temperature that controls the scaling.
Calibration performance can be enhanced by tuning $T$ on a held-out validation set.

\noindent\textbf{Focal Loss and Dual Focal Loss}
Some other previous works have attempted to improve model calibration by modifying the loss function.
Focal Loss~\citep{lin2017focal} was initially introduced to address the foreground-background imbalance problem in object detection.
It addresses the issue by incorporating a loss weighting factor based on sample complexity, reducing the weight of easy samples and allowing the model to focus on harder-to-classify instances.

Formally, given the predicted probability $\hat{p}(\vx)$ on sample $\vx$, the focal loss is defined as:
\begin{equation}
\mathcal{L}_{\text{FL}}(\vx,y)=-\sum_{i=1}^{K}y_i(1-\hat{p}_i)^\gamma\log \hat{p}_i(\vx),
\label{eq: focal loss}
\end{equation}
where $\gamma$ is a pre-defined hyperparameter.
Previous studies~\citep{mukhoti2020calibrating} have demonstrated that optimizing models using Focal Loss results in better calibration compared to Cross-Entropy.
This improvement is partly due to the entropy-based regularization effect introduced by Focal Loss, while complexity-based weighting also likely plays a role.
However, we noticed that Focal Loss was initially applied to binary classification, where the complexity term $(1-\hat{p}_i)$ is related to the Brier Score, which can also be used for evaluating uncertainty.
In multi-class classification, although complexity and uncertainty are not strictly equivalent, the complexity term still captures some class-level uncertainty.

Several variants of Focal Loss exist, including Dual Focal Loss~\citep{tao2023dual}, which incorporates the probability from the second most probable class into the scaling factor to address the under-confidence issue caused by Focal Loss:
\begin{equation}
\mathcal{L}_{\text{DFL}}(\vx,y)=-\sum_{i=1}^{K}y_i(1-\hat{p}_i(\vx)+\hat{p}_j(\vx))^\gamma\log \hat{p}_i(\vx),
\label{eq: dual focal loss}
\end{equation}
where $\hat{p}_j(x) = \max_i\{\hat{p}_i(x)\vert\hat{p}_i(x)<\hat{p}_{gt}(x)\}$.
\section{Method}
\label{sec:method}

\subsection{Weighting with Sample-wise Uncertainty}
\label{subsec: Weighting with sample-wise Uncertainty}

Focal Loss and Dual Focal Loss both support the concept that uncertainty-based weighting in conjunction with Cross Entropy can enhance model calibration.
We propose a generalized loss function framework, termed Uncertainty CE Loss, which incorporates a sample-wise uncertainty metric into Cross Entropy as a weighting factor:
\begin{equation}
    \mathcal{L}_{\text{\tiny Uncertainty}}(\vx,y)=-\sum_{i=1}^K u(\hat{p}_i(\vx))\cdot y_i\log\hat{p}_i(\vx),
    \label{eq: uncertainty loss}
\end{equation}
where $u(\hat{p}_i)$ is the adaptive term that evaluates the uncertainty of sample $\vx$.
The motivation behind this design is to use the predicted calibration error for each sample to scale the loss function, thereby directing the model's optimization more effectively towards samples with higher uncertainty.

Therefore, the weights used in Focal Loss is given by:
\begin{equation}
    u_{\text{\tiny FL}}(\hat{p}(\vx)) = (1-\hat{p}_c)^\gamma.
    \label{eq: weight for FL}
\end{equation}
And the scaling term in Dual Focal Loss is defined as:
\begin{equation}
    u_{\text{\tiny DFL}}(\hat{p}(\vx)) = (1-\hat{p}_c(x)+\hat{p}_j(x))^\gamma,
    \label{eq: weight for DFL}
\end{equation}
where $\hat{p}_j(x)$ is the second maximum confidence in prediction.
The weight used in Focal Loss can be seen as derived from the Brier Score, focusing on the error of the ground true class.
In contrast, the scaling term in Dual Focal Loss incorporates uncertainty information from the second most probable class.
However, directly weighting loss function with certain uncertainty metrics, as those used in Focal Loss and Dual Focal Loss, presents several challenges.

Reviewing our objectives, we aim to weight the loss function value by each sample's uncertainty, resulting in greater optimization steps for those samples with higher uncertainty.
Our focus actually is not on the loss value itself, but on the optimization.
The purpose of uncertainty weighting is to focus the model's attention on samples with higher uncertainty.
Thus, the key factor is the gradient: higher weights should lead to larger gradients, resulting in more substantial model updates for those samples.
For Cross Entropy, its value is positively correlated with the gradient: a higher CE value yields a larger gradient, leading to more significant updates for optimization.
When applying a scalar weight to CE, this relationship remains intact, as it does not disrupt the positive correlation.

\begin{figure}[t]
    \centering
    \includegraphics[width=1.0\columnwidth]{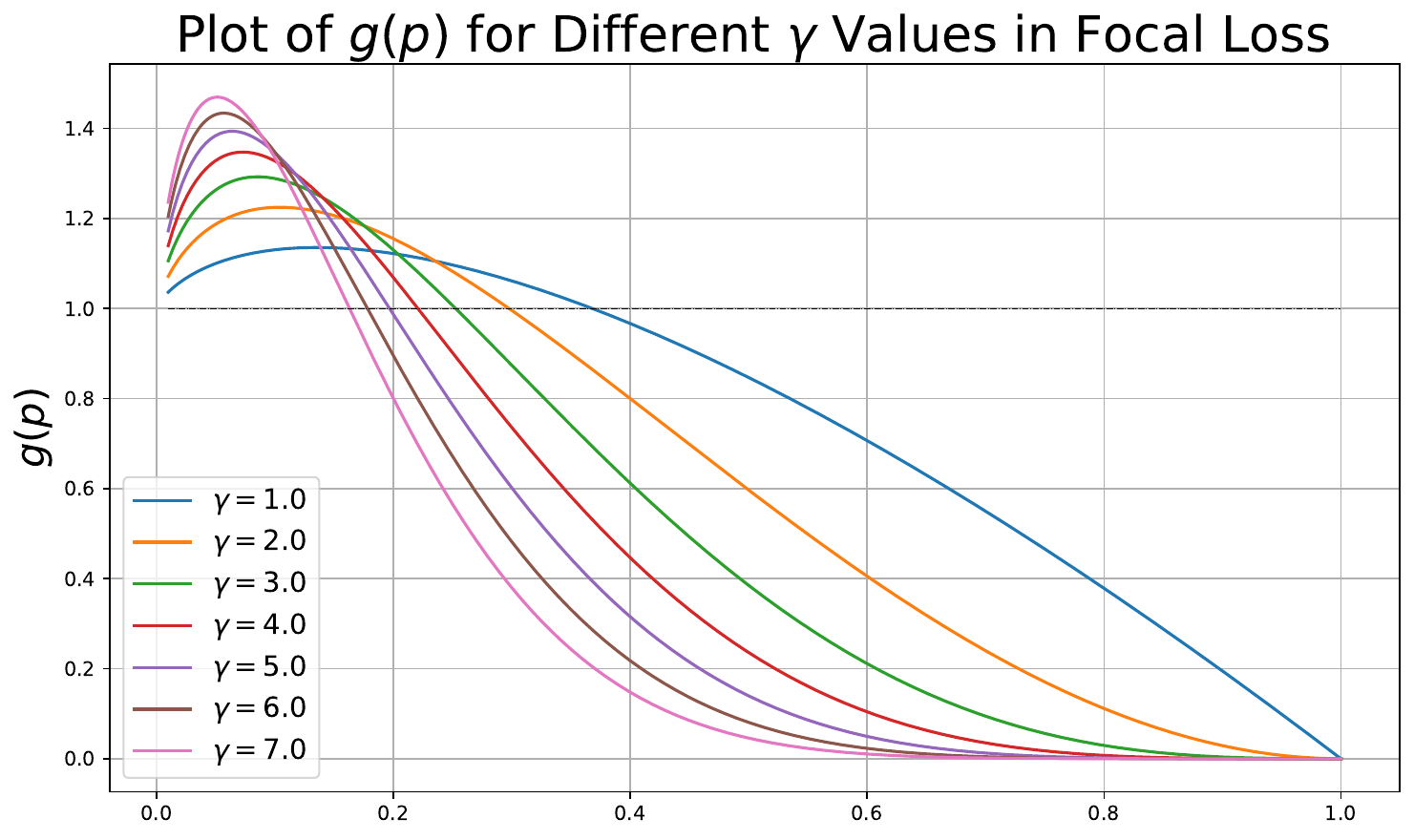}
    \caption{$g(p,\gamma)$ of Focal Loss vs predicted confidence $\hat{p}_c$.}
    \label{fig: grad of focal loss vs confidence}
\end{figure}
However, for differentiable weights, directly applying them to the loss can disrupt the positive correlation between the loss and the gradient, ultimately impairing effective model optimization.
To further analyze this, let us consider Focal Loss.
For Focal Loss $\mathcal{L}_{\text{\tiny FL}}$ and Cross Entropy $\mathcal{L}_{\text{\tiny CE}}$, their gradients are given by: $\frac{\partial}{\partial\vw}\mathcal{L}_{\text{\tiny FL}}=g(\hat{p}_i,\gamma)\frac{\partial}{\partial\vw}\mathcal{L}_{\text{\tiny CE}}$, where $g(p,\gamma) = (1 - p)^\gamma - \gamma p(1 - p)^{\gamma - 1}\log(p)$, $\gamma$ is a predefined hyperparameter, and $\vw$ represents the parameters of the final linear layer.
From a gradient perspective, $g(p,\gamma)$ acts as a weight on the CE gradient, which is illustrated in Figure~\ref{fig: grad of focal loss vs confidence}.
There exists a point $p_0\in[0,1]$ such that within the range $[0,p_0]$, $\frac{\partial}{\partial p}g(p,\gamma)>0$ and within the range $[p_0,1]$, $\frac{\partial}{\partial p}g(p,\gamma)<0$.
In binary classification, $1-\hat{p}_i$ is related to the Brier Score that reveals calibrated error, indicating that the predicted calibrated error is not always positively correlated with the weight on gradient.
As uncertainty decreases, the weight on gradient initially increases, before eventually aligning with the level of uncertainty.
This causes the model to focus more on samples with moderate uncertainty rather than those with the highest uncertainty, which contradicts our original goal of emphasizing the most uncertain samples.
Further, changes in uncertainty are not immediately reflected in the sample weights; instead, multiple training iterations are required for these changes to align properly.
Thus, directly applying some kind of uncertainty metrics into Eq.~\ref{eq: uncertainty loss} would pose issues to the uncertainty weighting.

\subsection{Sample-wise Uncertainty on Gradients}
\label{subsec: Weighting Sample-wise Uncertainty on Gradients}

The discussion in Sec.~\ref{subsec: Weighting with sample-wise Uncertainty} motivates us to ensure the weight on gradients for a sample to align with its uncertainty.
This could be intuitively solved by directly applying the model uncertainty on samples as a weight of the gradient.
We define the modified gradient as:
\begin{equation}
    \frac{\partial}{\partial \theta}\mathcal{L}_{\text{\tiny Uncertainty-GRA}}(\vx,y)=u(\hat{p}(\vx))\frac{\partial}{\partial \theta}\mathcal{L}_{\text{\tiny CE}}(\vx,y).
    \label{eq: uncertainty-gra loss gradient}
\end{equation}
Taking a SGD optimizer as an example, the model update can be expressed as: $\theta_{t+1}=\theta+\alpha\cdot\frac{\partial}{\partial \theta}\mathcal{L}_{\text{\tiny Uncertainty-GRA}}(\vx,y)$,
where $\theta$ is the model parameters and $\alpha$ is the learning rate.
When the predicted uncertainty on sample $\vx$ is higher, the optimization step results in a larger update, encouraging the model to focus more on uncertain samples.
A general form of the loss function for our uncertainty-weighted gradient framework, termed Uncertainty-GRA CE Loss, is defined as:
\begin{equation}
    \mathcal{L}_{\text{\tiny Uncertainty-GRA}}(\vx,y)=-\int\sum_{i=1}^K u(\hat{p}(\vx))\cdot \frac{y_i}{\hat{p}_i(\vx)}\text{d}\hat{p}(\vx),
    \label{eq: uncertainty loss on gra}
\end{equation}
In practice, to compute the $\mathcal{L}_{\text{\tiny Uncertainty-GRA}}$, we can detach the gradient of $u(\hat{p}(\vx))$ and multiply it with the Cross Entropy, instead of calculating the gradient integration.

\begin{figure*}[t]
    \centering
    \begin{subfigure}[b]{0.3\textwidth}
        \centering
        \includegraphics[width=\textwidth]{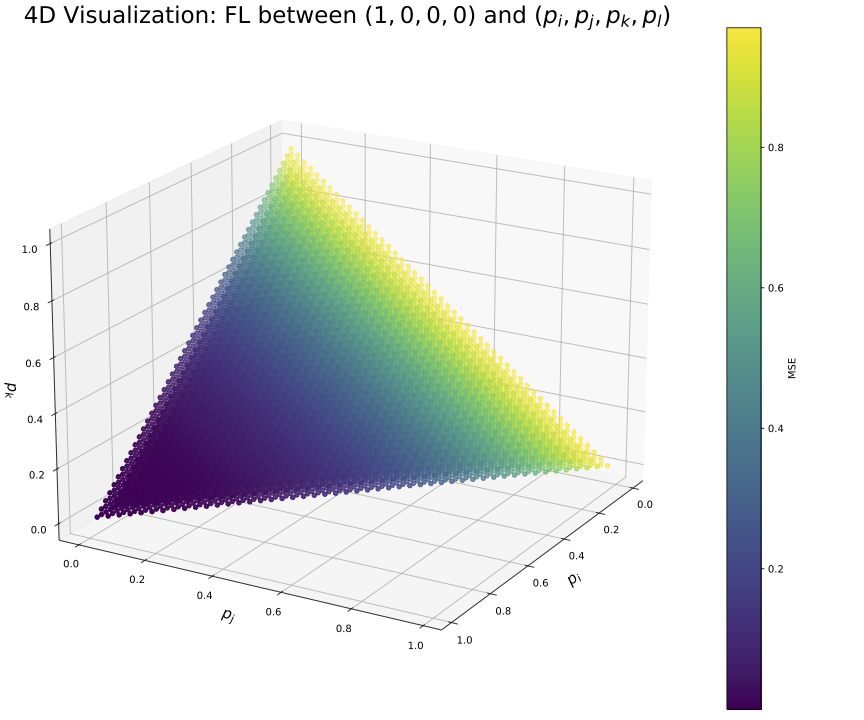}
        \caption{An illustration of the $u_{\text{\tiny FL}}$.}
        \label{fig: dataset}
    \end{subfigure}
    \hfill
    \begin{subfigure}[b]{0.3\textwidth}
        \centering
        \includegraphics[width=\textwidth]{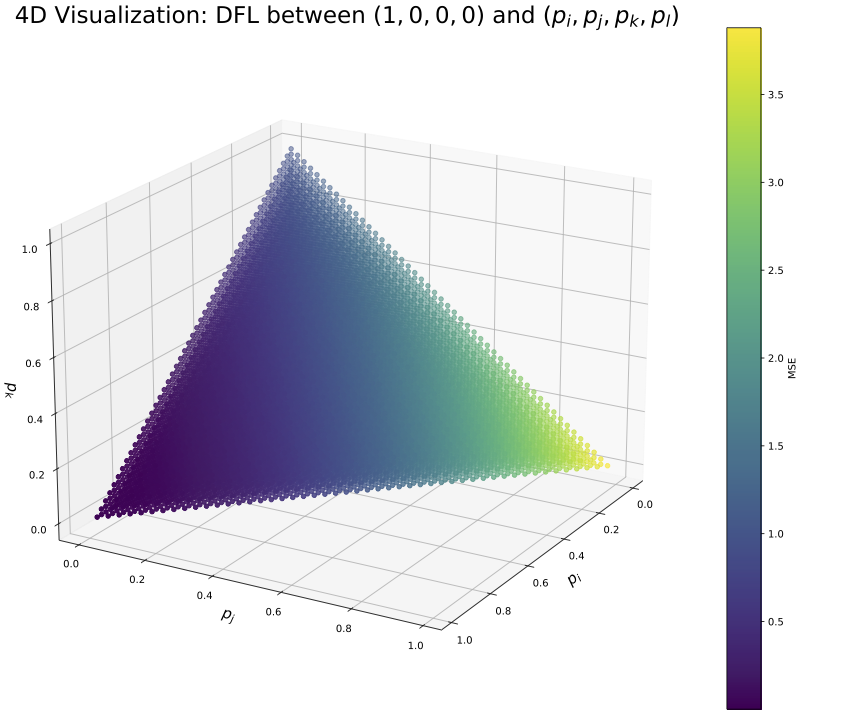}
        \caption{An illustration of the $u_{\text{\tiny DFL}}$.}
        \label{fig: reliability diagram for CE}
    \end{subfigure}
    \hfill
    \begin{subfigure}[b]{0.3\textwidth}
        \centering
        \includegraphics[width=\textwidth]{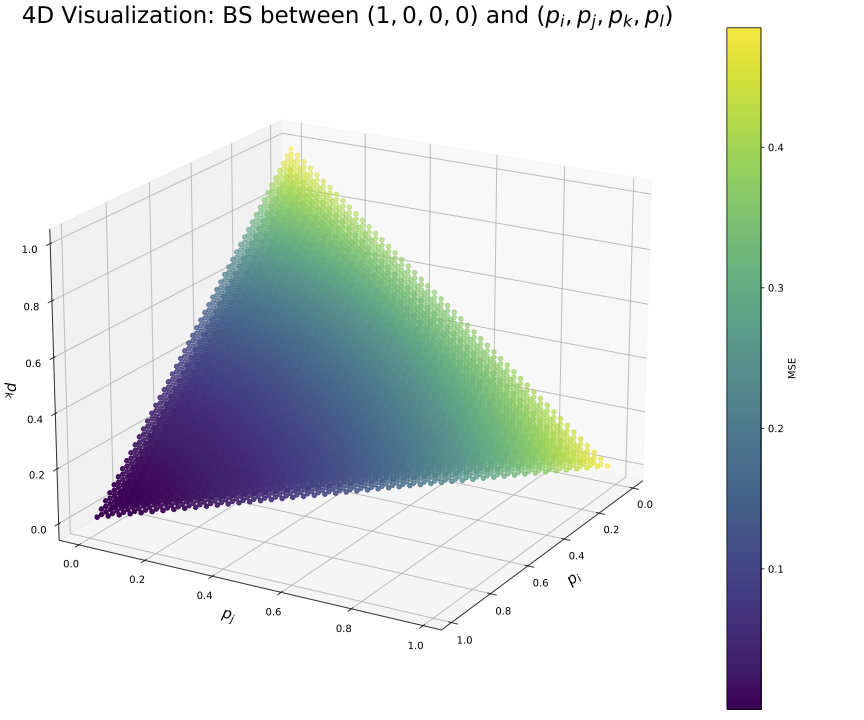}
        \caption{An illustration of the $u_{\text{\tiny BS}}$.}
        \label{fig: reliability diagram for calibrated error CE}
    \end{subfigure}
    \caption{An illustration of value of gradient weight function on a 4 class classification. It is obvious that $u_{\text{\tiny FL}}$ varies only along the $p_i$ axis and $u_{\text{\tiny DFL}}$ changes along the $p_i$ and $p_j$ axes. $u_{\text{\tiny BS}}$ responds to changes across all axes, providing a more complete uncertainty evaluation.}
    \label{fig: 4d illustration}
\end{figure*}

\subsection{Sample-wise Uncertainty Metrics}
\label{subsec: Weighting with sample-wise uncertainty metrics}

With Eq.~\ref{eq: uncertainty loss on gra}, calibration can be achieved by employing the sample's ground-truth uncertainty as the scaling term.
However, accessing the ground-truth uncertainty for real-world image samples is impractical.
Therefore, alternative metrics are required to estimate the sample-wise uncertainty $u(\vx)$.

One approach to evaluate sample-wise uncertainty is based on computing the generalized Brier Score (gBS):
\begin{equation}
    u_{\text{\tiny gBS}}(\hat{p}(\vx)) = \sum_{i=1}^K \Vert\hat{p}_i(\vx)-y_i\Vert^\gamma_\beta.
    \label{eq: weight by BS}
\end{equation}
We provide a brief discussion on the effectiveness of the generalized Brier Score (gBS) as a measure of calibration error.
When using the Brier Score (BS) to evaluate calibration error, with $\beta=2$ and $\gamma=2$, the difference between the expected and predicted calibration error is given by:
\begin{align}
    c(\vx)-u_{\text{\tiny BS}}(\hat{p}(\vx))&=\mathbb{E}_{y\sim\eta(x)}[\Vert\hat{p}(\vx)-\rveta(\vx)\Vert^2_2-\Vert\hat{p}(\vx)-y\Vert^2_2]\notag \\ 
                       &=\sum^K_{k=1}\rveta_k(\vx)(\rveta_k(\vx)-1).
    \label{eq: effectiveness of BS}
\end{align}
Thus, the term $\sum^K_{k=1}\rveta_k(\vx)(\rveta_k(\vx)-1)$ depends solely on $\rveta(\vx)$ which is fixed for a given sample $\vx$.
Consequently, the error remains constant for a specific sample $\vx$.
We provide a detailed computation of Eq.~\ref{eq: effectiveness of BS} in Appendix.

Several variants of $u_{\text{\tiny gBS}}(\hat{p})$ have demonstrated their effectiveness in previous studies.
When $\beta=1$ and only the actual class is considered, the generalized Brier Score can be interpreted as the scaling term in Focal Loss in Eq.~\ref{eq: weight for FL}.
When $\beta=1$ and both the actual class and the maximum predicted class (excluding the actual class) are considered, the generalized Brier Score can be interpreted as the scaling term in Dual Focal Loss in Eq.~\ref{eq: weight for DFL}.

However, these two metrics only consider several classes in the predicted probability .
In the case of Focal Loss, which was originally designed for binary classification, considering one-dimensional uncertainty is sufficient due to the constraint that the sum of probabilities must equal 1.
When extended to multi-class classification, these metrics are not accurate enough to evaluate uncertainty.

We visualize the gradient weights $u_{\text{\tiny FL}}$, $u_{\text{\tiny DFL}}$ and $u_{\text{\tiny BS}}$ in Figure~\ref{fig: 4d illustration} for a 4-class scenario.
The three axes represent three dimensions of the predicted probability, while the fourth is implied by the probability sum constraint.
From the figure, it is evident that $u_{\text{\tiny FL}}$ and $u_{\text{\tiny DFL}}$ are sensitive only to changes in one or two dimensions.
However, different points in the coordinate system have different uncertainties.
They fail to adequately capture uncertainty change across all dimensions.
But the $u_{\text{\tiny BS}}$ can accurately respond to changes along any coordinate axis.
\begin{table*}
  \centering
  \resizebox{\textwidth}{!}{%
  \begin{tabular}{l c c c c c c c c c c c c c c c c c}
    \toprule
    \multirow{2}{*}{Dataset} & \multirow{2}{*}{Model} 
    & \multicolumn{2}{c}{CE} & \multicolumn{2}{c}{BL} & \multicolumn{2}{c}{MMCE} & \multicolumn{2}{c}{FLSD} & \multicolumn{2}{c}{DFL} & \multicolumn{2}{c}{BSCE} & \multicolumn{2}{c}{BSCE-GRA}\\
    & & Pre T & Post T & Pre T & Post T & Pre T & Post T & Pre T & Post T & Pre T & Post T & Pre T & Post T & Pre T & Post T \\
    \midrule
    \multirow{4}{*}{CIFAR10}
    & ResNet50   & 4.36 & 1.32 & 4.29 & 1.50 & 4.48 & 1.41 & 1.26 & 1.15 & 1.00 & 1.00 & 0.88 & 0.88 & \textbf{0.74} & \textbf{0.74} \\
    & ResNet110  & 4.7  & 1.56 & 4.48 & 1.66 & 4.80 & 1.29 & 1.81 & 1.17 & 1.01 & 1.01 & 0.99 & 0.99 & \textbf{0.87} & \textbf{0.87} \\
    & WideResNet & 3.35 & \textbf{0.94} & 2.86 & 1.10 & 3.65 & 1.28 & 1.84 & 1.04 & 3.32 & 1.16 & 1.7  & 0.95 & \textbf{1.46} & 1.12 \\
    & DenseNet   & 4.64 & 1.46 & 3.96 & 1.46 & 4.81 & 1.67 & 1.37 & 1.17 & 0.87 & \textbf{0.77} & 1.01 & 1.01 & \textbf{0.87} & 0.87 \\
    \midrule
    \multirow{4}{*}{CIFAR100}
    & ResNet50   & 18.05 & 3.05 & 7.87 & 4.27 & 15.87 & 3.32 & 5.53 & 2.57 & 2.54 & 2.56 & 1.90 & 1.90 & \textbf{1.59} & \textbf{1.59} \\
    & ResNet110  & 18.84 & 4.63 & 16.77 & 4.30 & 18.65 & 3.93 & 6.88 & 3.71 & 3.47 & 3.47 & 2.75 & 2.75 & \textbf{2.53} & \textbf{2.53} \\
    & WideResNet & 14.81 & 3.27 & 7.74 & 4.46 & 14.58 & 2.99 & 2.70 & 2.71 & 5.45 & 2.52 & 2.63 & \textbf{2.42} & \textbf{2.46} & 2.46 \\
    & DenseNet   & 19.1  & 3.43 & 8.13 & 2.99 & 17.56 & 2.87 & 3.38 & 1.30 & 4.68 & 1.83 & 1.63 & 1.62 & \textbf{1.62} & \textbf{1.61} \\
    \midrule
    \multirow{1}{*}{TinyImageNet}
    & ResNet50   & 14.94 & 5.16 & 7.81 & 1.47 & 14.58 & 2.99 & \textbf{2.18} & 2.18 & 6.71 & 2.28 & 4.0 & 1.76 & 4.57 & \textbf{1.47} \\
    \bottomrule
  \end{tabular}%
  }
  \caption{\textbf{Comparison of Calibration Methods Using ECE Across Various Datasets and Models.} ECE values are reported using 15 bins, with the best-performing method for each dataset-model combination highlighted in bold. Results are averaged over three runs with different random seeds.}
  \label{tab: ECE results of experiment}
\end{table*}

\begin{figure*}[t]
    \centering
    \begin{subfigure}[b]{0.16\textwidth}
        \centering
        \includegraphics[width=\textwidth]{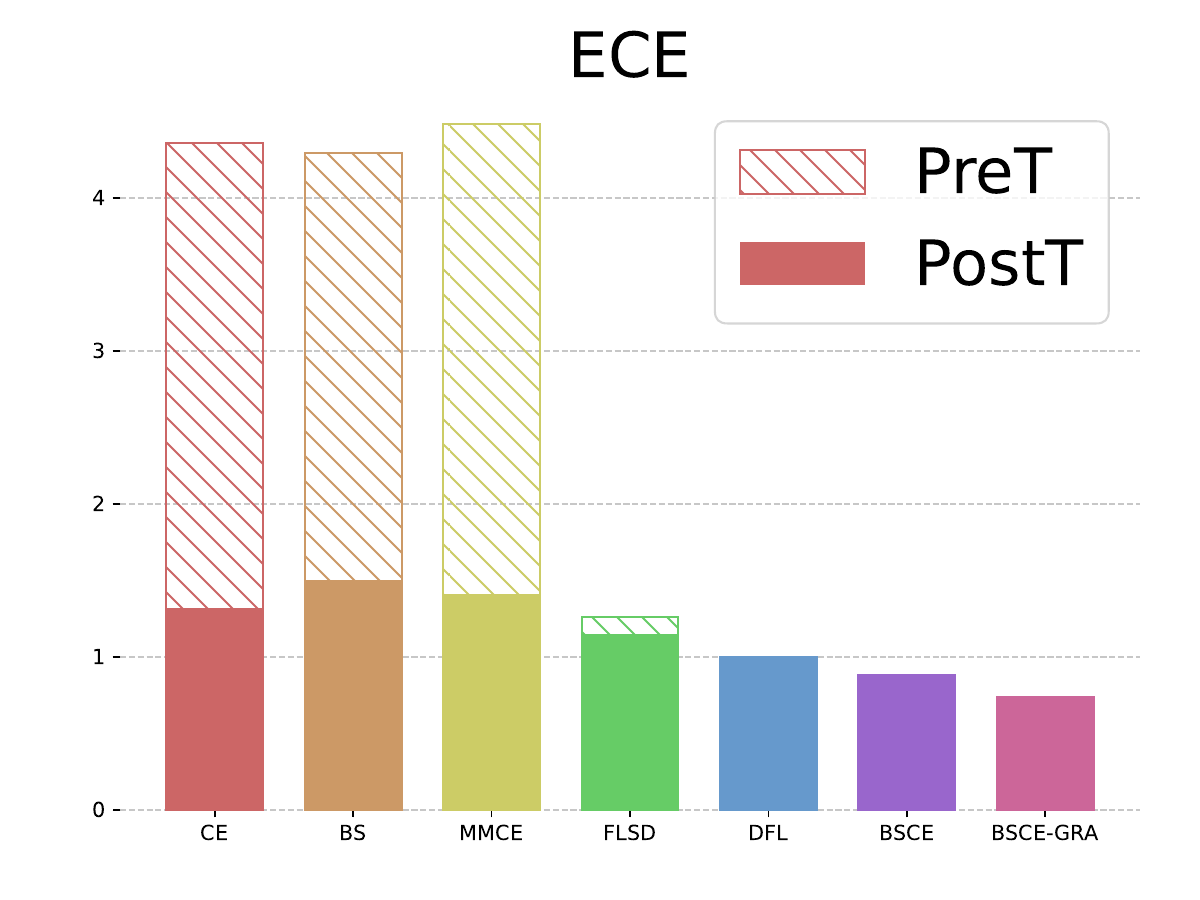}
        \end{subfigure}
    \hfill
    \begin{subfigure}[b]{0.16\textwidth}
        \centering
        \includegraphics[width=\textwidth]{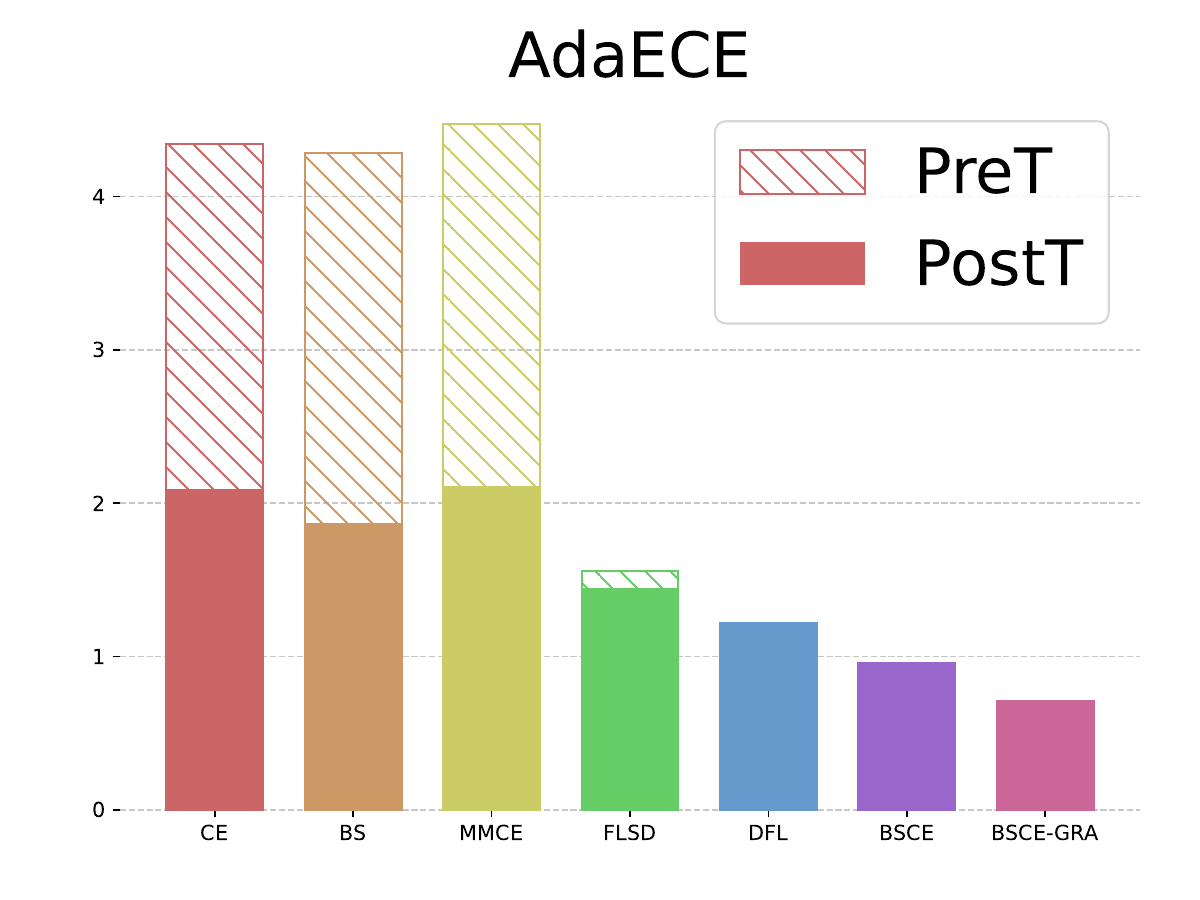}
    \end{subfigure}
    \hfill
    \begin{subfigure}[b]{0.16\textwidth}
        \centering
        \includegraphics[width=\textwidth]{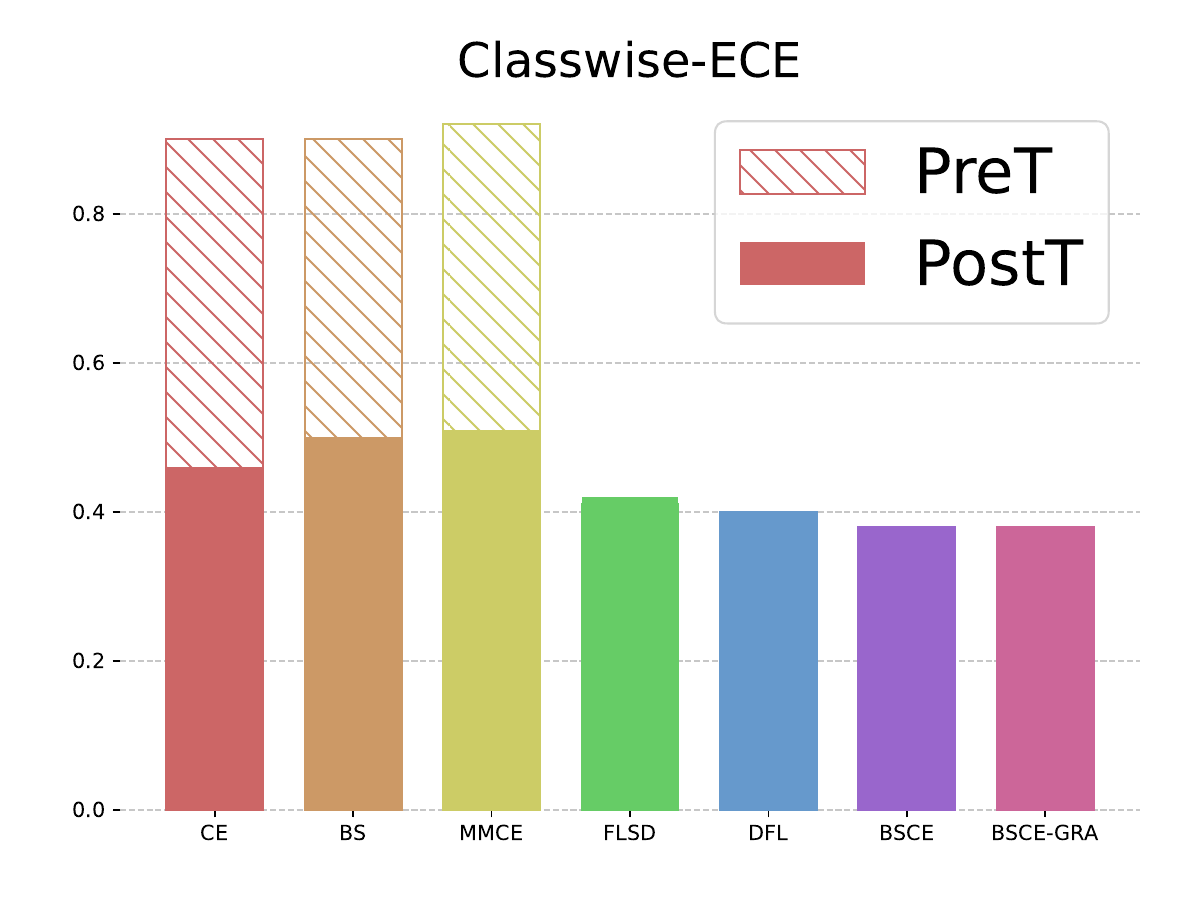}
    \end{subfigure}
    \hfill
    \begin{subfigure}[b]{0.16\textwidth}
        \centering
        \includegraphics[width=\textwidth]{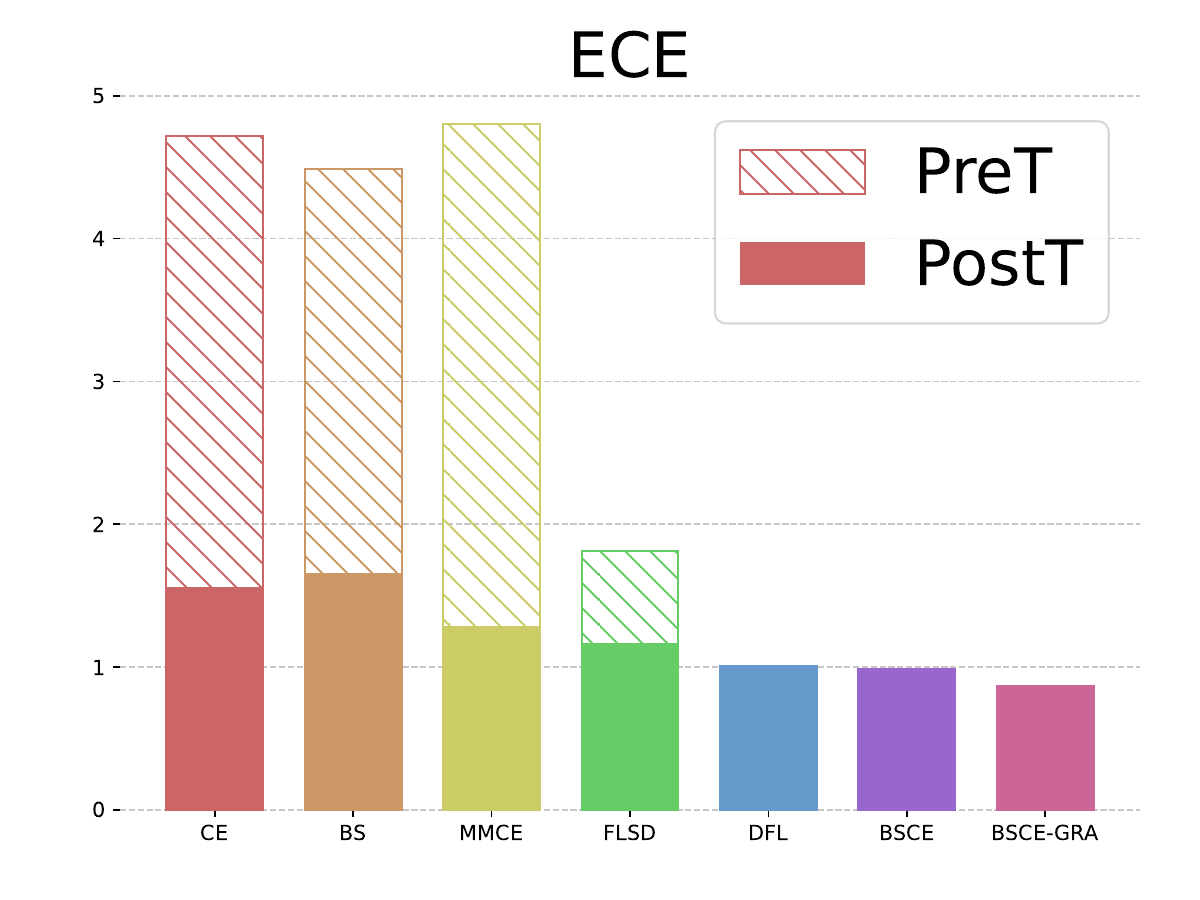}
    \end{subfigure}
    \hfill
    \begin{subfigure}[b]{0.16\textwidth}
        \centering
        \includegraphics[width=\textwidth]{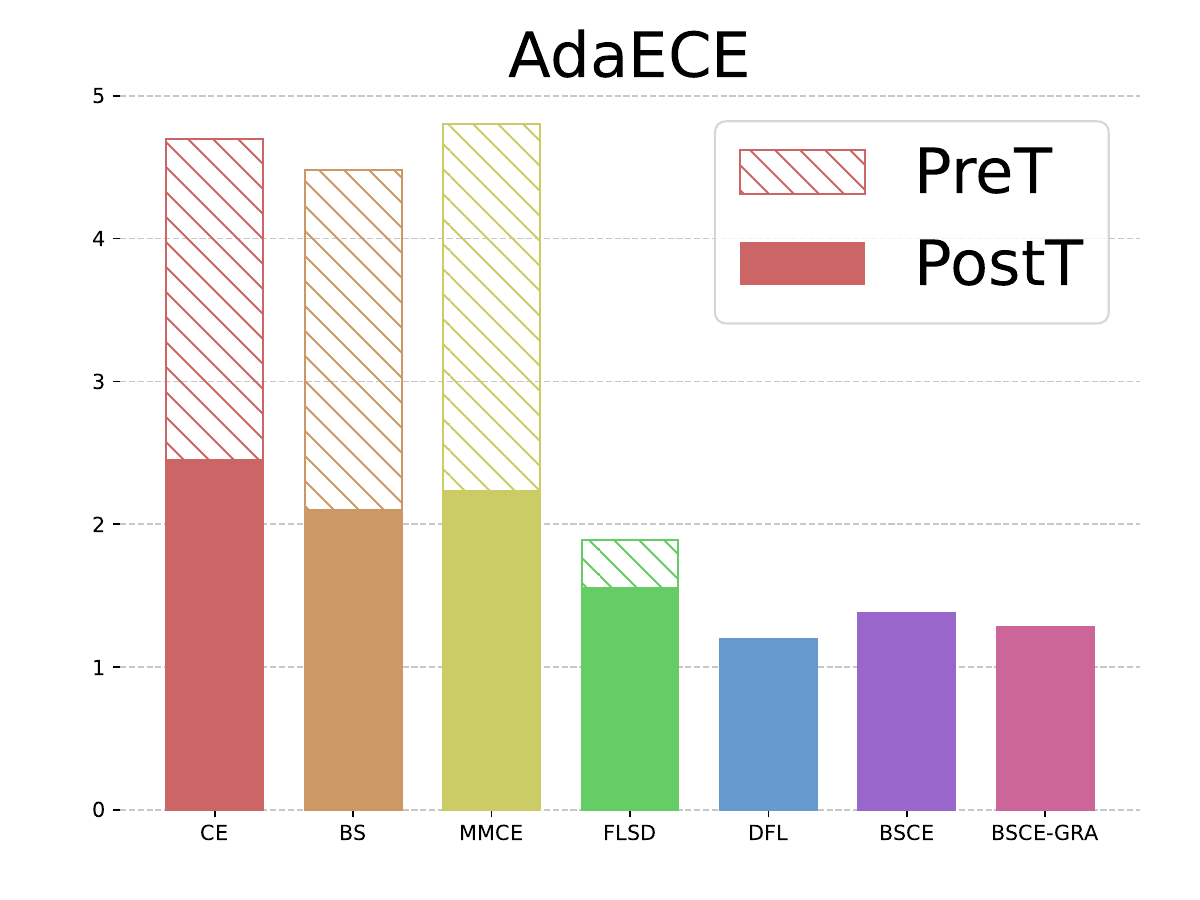}
    \end{subfigure}
    \hfill
    \begin{subfigure}[b]{0.16\textwidth}
        \centering
        \includegraphics[width=\textwidth]{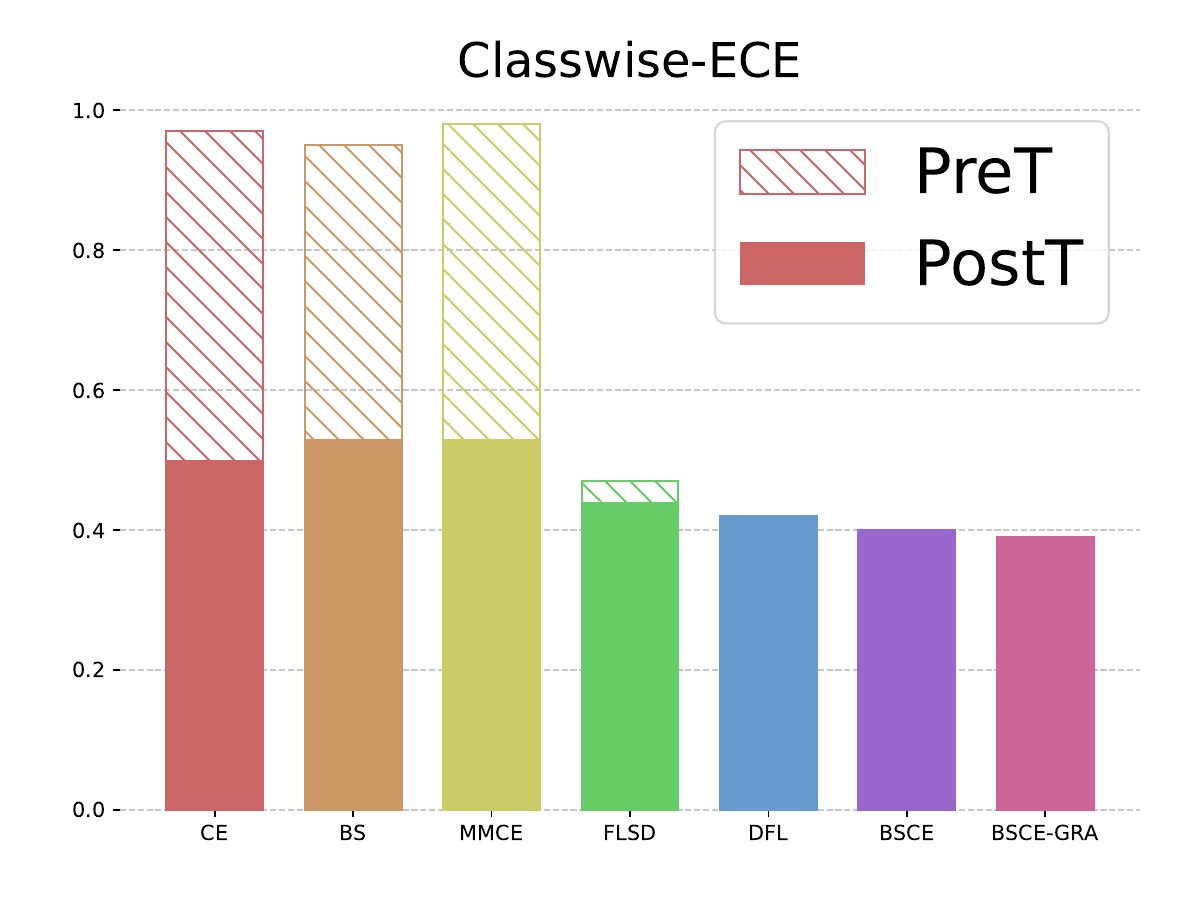}
    \end{subfigure}
    \caption{Comparison of different ECE metrics. The first three plots show the uncertainty for CIFAR-10 using ResNet-50, while the remaining plots represent ResNet-110 on CIFAR-10.}
    \label{fig: bar plot of ECEs}
\end{figure*}

To further evaluate whether these metrics accurately measure the ground truth uncertainty of samples, we conducted experiments on a toy dataset.
The toy dataset consists of 5 two-dimensional Gaussian distributions, representing 5 groups of data: $\gN(\mu_i,\Sigma),i\in\{0,...,5\}$.
The mean vectors $\mu_i$ were randomly sampled from the range $[-10,10]$ and $\Sigma=\mI$ was used as the shared covariance matrix for all groups.
We generated 10,000 data points from each group to form the training dataset, which was used to train a two layer CNN model for 5 epochs.
An additional 1,000 samples from each distribution were used to create the test dataset.
Therefore, the ground truth probability $\eta(\vx)$ can be computed using the probability density function (PDF) of each distribution:
\begin{equation}
    \eta(\vx)=\frac{p^{n}(\vx)}{\sum_i^{N} p^{i}(\vx)},
    \label{eq: ground truth probability}
\end{equation}
where $N=5$ in this case, $p^{n}(\vx)$ and $p^{i}(\vx)$ are the PDF of corresponds class for $\vx$ and remaining classes, respectively.
The ground truth uncertainty of each sample is calculated by Eq.~\ref{eq: calibrated error}, representing the model's ground truth uncertainty for each sample, along with the sample-wise uncertainty metrics $u_{\text{\tiny FL}}$, $u_{\text{\tiny DFL}}$ and $u_{\text{\tiny gBS}}$.
To validate the accuracy of these metrics, we used Pearson correlations between the calibrated error and the uncertainty metrics to verify their positive correlation among 5 runs.
We use grid-search to find the optimal hyper-parameter for each methods.
Among the methods, gBS achieves the highest Pearson correlation coefficient of 0.664, indicating the strongest linear relationship between the predicted values and the true targets.
DFL follows with a correlation of 0.638, while FL has the lowest correlation at 0.550.
This suggests that gBS provides the most accurate predicted uncertainty alignment with ground-truth uncertainty, followed by DFL and FL.

Based on the results, we conclude that although ground-truth uncertainty is inaccessible for real-world datasets, these alternative metrics provide a reliable means of estimating uncertainty, making them suitable for use during training.
Besides, the metric $u_{\text{\tiny gBS}}$ has the best performance compared to the uncertainty metric used in Focal Loss and Dual Focal Loss.
We incorporate $u_{\text{\tiny gBS}}$ into the gradient-weighted framework discussed in Section~\ref{subsec: Weighting Sample-wise Uncertainty on Gradients} and introduce a new loss function called BSCE-GRA. This loss function uses the generalized Brier Score as an adaptive uncertainty metric to weight the Cross Entropy gradients, defined as:
\begin{equation}
    \mathcal{L}_{\text{\tiny BSCE-GRA}}(\vx,y)=-\int\sum_{i=1}^K u_{\text{\tiny gBS}}(\hat{p}(\vx))\cdot \frac{y_i}{\hat{p}_i(\vx)}\text{d}\hat{p}(\vx).
    \label{eq: BSCE-GRA}
\end{equation}
To further validate the effectiveness of proposed method, we provide a comprehensive theoretical evidence that optimizing with BSCE-GRA, the K-class predicted probability $q$
would equal the actual class-posterior probability $\eta$, thereby preventing over/under-confidence when convergence.
Due to page limitation, the proof is provided in Appendix.

Besides, the Uncertainty-GRA Loss makes it possible to use non-differentiable uncertainty metrics for model calibration.
However, it requires the weights to capture sample-wise uncertainty, whereas some existing measurements like ECE can only compute the uncertainty for a group of samples, making them unsuitable for the proposed framework.
\begin{figure*}[t]
    \centering
    \begin{subfigure}[b]{0.23\textwidth}
        \centering
        \includegraphics[width=\textwidth]{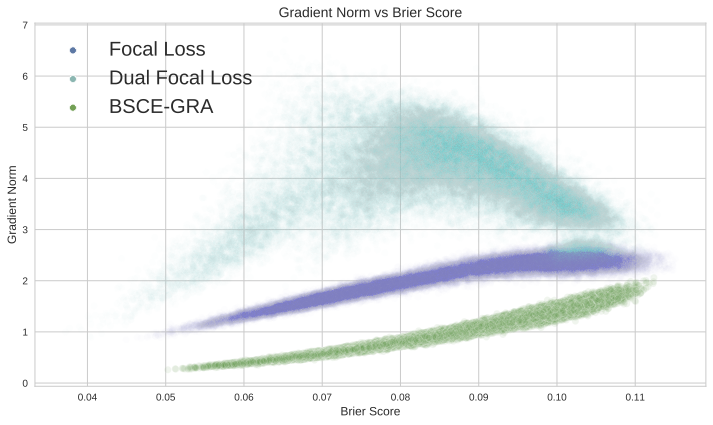}
        \caption{Gradient norm distribution at epoch 50}
    \end{subfigure}
    \hfill
    \begin{subfigure}[b]{0.23\textwidth}
        \centering
        \includegraphics[width=\textwidth]{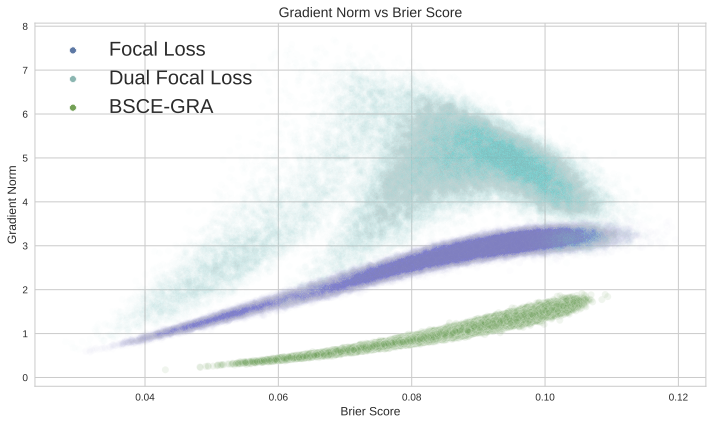}
        \caption{Gradient norm distribution at epoch 150}
    \end{subfigure}
    \hfill
    \begin{subfigure}[b]{0.23\textwidth}
        \centering
        \includegraphics[width=\textwidth]{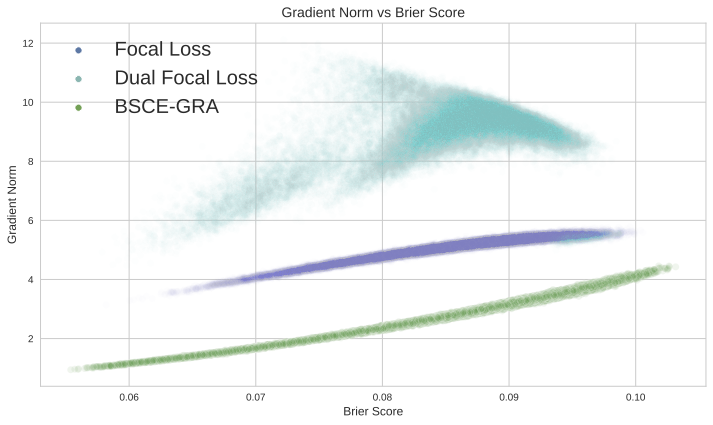}
        \caption{Gradient norm distribution at epoch 250}
    \end{subfigure}
    \hfill
    \begin{subfigure}[b]{0.23\textwidth}
        \centering
        \includegraphics[width=\textwidth]{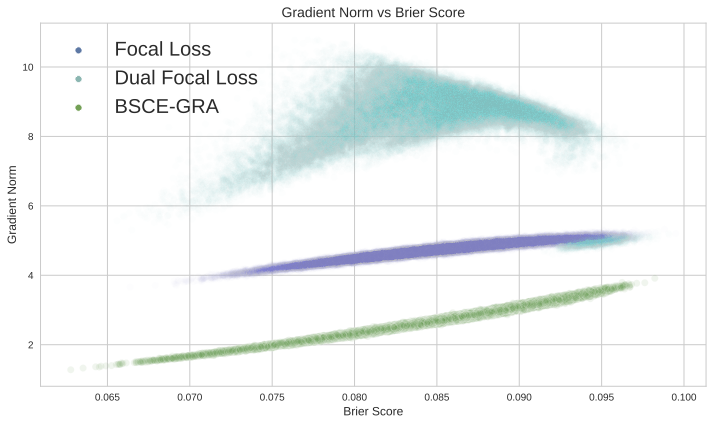}
        \caption{Gradient norm distribution at epoch 350}
    \end{subfigure}
    \caption{Evolution of gradient norm distributions across different training epochs for various loss functions. The scatter plots show the relationship between gradient norm and Brier Score for different loss functions (Focal Loss, Dual Focal Loss, BSCE-GRA).}
    \label{fig: gradient bs}
\end{figure*}

\begin{table*}
  \centering
  \resizebox{\textwidth}{!}{%
  \begin{tabular}{l cc cc cc cc cc cc}
    \toprule
    Metrics & \multicolumn{2}{c}{FLSD} & \multicolumn{2}{c}{FLSD-GRA} & \multicolumn{2}{c}{DFL} & \multicolumn{2}{c}{DFL-GRA} & \multicolumn{2}{c}{BSCE} & \multicolumn{2}{c}{BSCE-GRA} \\
    & pre T & post T & pre T & post T & pre T & post T & pre T & post T & pre T & post T & pre T & post T \\
    \midrule
    Acc & \textbf{95.04\%} & \textbf{95.04\%} & 94.72\% & 94.72\% & 94.63\% & 94.63\% & 94.76\% & 94.76\% & 95.03\% & 95.03\% & 94.69\% & 94.69\% \\
    ECE & 1.26 & 1.15 & 0.88 & 0.88 & 1.00 & 1.00 & 0.93 & 0.84 & 0.88 & 0.88 & \textbf{0.74} & \textbf{0.74} \\
    Ada ECE & 1.56 & 1.45 & 1.19 & 1.19 & 1.22 & 1.22 & 0.77 & 0.81 & 0.96 & 0.96 & \textbf{0.71} & \textbf{0.71} \\
    % Classwise ECE & 0.41 & 0.42 & 0.40 & 0.40 & 0.40 & 0.40 & 0.40 & 0.38 & \textbf{0.38} & \textbf{0.38} & 0.40 & 0.40 \\
    % Brier Score & 7.81 & 7.78 & 7.99 & 7.99 & 8.23 & 8.23 & 8.08 & 8.09 & \textbf{7.66} & \textbf{7.66} & 8.18 & 8.18 \\
    \bottomrule
  \end{tabular}%
  }
  \caption{Comparison of weighting different uncertainty metrics on gradient or loss function, including $u_{\text{\tiny FL}}$ and $u_{\text{\tiny FL}}$. The results validate the effectiveness of the gradient-weighting strategy among different uncertainty metrics.}
  \label{tab: gradient or not experiment}
\end{table*}

\section{Experiment}
\label{sec: experiment}

We evaluate our methods on multiple deep neural networks (DNNs), including ResNet50, ResNet110~\citep{he2016deep}, WideResNet~\citep{zagoruyko2016wide}, and DenseNet~\citep{huang2017densely}.
Our experiments are conducted on CIFAR-10, CIFAR-100~\citep{krizhevsky2009learning}, and Tiny-ImageNet~\citep{deng2009imagenet} to assess calibration performance.
Further details about the datasets can be found in the appendix.

\noindent\textbf{Baselines.}
We compare our methods, BSCE, BSCE-GRA, and ECE-CE, with multiple existing approaches, including training with Cross Entropy (CE), Brier Loss (BL)\citep{brier1950verification}, MMCE Loss\citep{kumar2018trainable}, Focal Loss with Adaptive Exponent (FLSD)\citep{mukhoti2020calibrating}, and Dual Focal Loss\citep{tao2023dual}.
For Focal Loss, we employ the FLSD-53 strategy~\citep{mukhoti2020calibrating} to adaptively adjust the gamma value sample-wise, setting $\gamma_{\text{\tiny FL}}=5$ for $\hat{p}_c\in[0,0.2)$ and $\gamma_{\text{\tiny FL}}=3$ for $\hat{p}_c\in[0.2,1)$.
For Dual Focal Loss, the gamma value is set to 5, as reported in the original work.

\noindent\textbf{Training Setup.}
Our training setup follows prior works~\citep{mukhoti2020calibrating, tao2023dual}.
We train CIFAR-10 and CIFAR-100 for 350 epochs, using 5,000 images from the training set for validation.
The learning rate is initially set to 0.1 for the first 150 epochs, then reduced to 0.01 for the next 100 epochs, and further reduced to 0.001 for the remaining epochs.
For Tiny-ImageNet, we train for 100 epochs, with the learning rate set to 0.1 for the first 40 epochs, 0.01 for the next 20 epochs, and 0.001 for the remaining epochs.
All experiments are conducted using SGD with a weight decay of $5\times 10^{-4}$ and a momentum of 0.9. The training and testing batch sizes for all datasets are set to 128.
We re-run all baseline methods using three different random seeds (1, 42, and 71), and report the average results.
All experiments are performed on a single Nvidia 4090 GPU.
For temperature scaling, the temperature parameter $T$ is optimized through a grid search with $T\in[0,0.1,0.2,...,10]$ on the validation set, selecting the value that yields the best post-temperature-scaling Expected Calibration Error (ECE).
The same optimized temperature parameter is applied to other metrics, such as AdaECE.
Further details on the datasets and additional experimental setup are provided in the appendix.

\subsection{Calibration Performance}

We report the average ECE before and after temperature scaling among three random seeds, along with the corresponding optimal temperatures, in Table~\ref{tab: ECE results of experiment}.
BSCE-GRA achieves state-of-the-art ECE performance in most cases, particularly in the pre-temperature-scaling results.
Notably, the fact that most of the optimal temperatures for BSCE-GRA are found to be 1 indicates that BSCE-GRA trains an inherently calibrated model, capable of achieving strong calibration performance without the need for additional temperature scaling.
This is a crucial advantage for developing accurate and reliable models that are efficient and require minimal post-processing.
The results on CIFAR-10 generally show better calibration performance compared to datasets with more labels (e.g., CIFAR-100 and Tiny-ImageNet) across multiple models.
Regarding network architecture, ResNet-50 demonstrates the best calibration performance among the four DNNs tested (ResNet-50, ResNet-110, Wide-ResNet-26-10, and DenseNet-121) on both CIFAR-10 and CIFAR-100 datasets.

\noindent\textbf{Different Metrics.}
The methods are further evaluated using several widely-accepted metrics to assess calibration performance across models, including Adaptive ECE and Classwise-ECE.
Adaptive ECE measures the expected calibration error while accounting for the distribution of the data, whereas Classwise-ECE is a variant that evaluates calibration error for each class individually.
Figure~\ref{fig: bar plot of ECEs} presents the results of multiple methods using ResNet-50 and ResNet-110 on the CIFAR-10 dataset.
The figure demonstrates that BSCE-GRA is the only method that achieves both inherently calibrated models and state-of-the-art performance across various metrics.
Additional results are presented in the appendix, providing further evidence of the effectiveness of our method in model calibration.

\begin{figure*}[t]
    \centering
    \begin{subfigure}[b]{0.3\textwidth}
        \centering
        \includegraphics[width=\columnwidth]{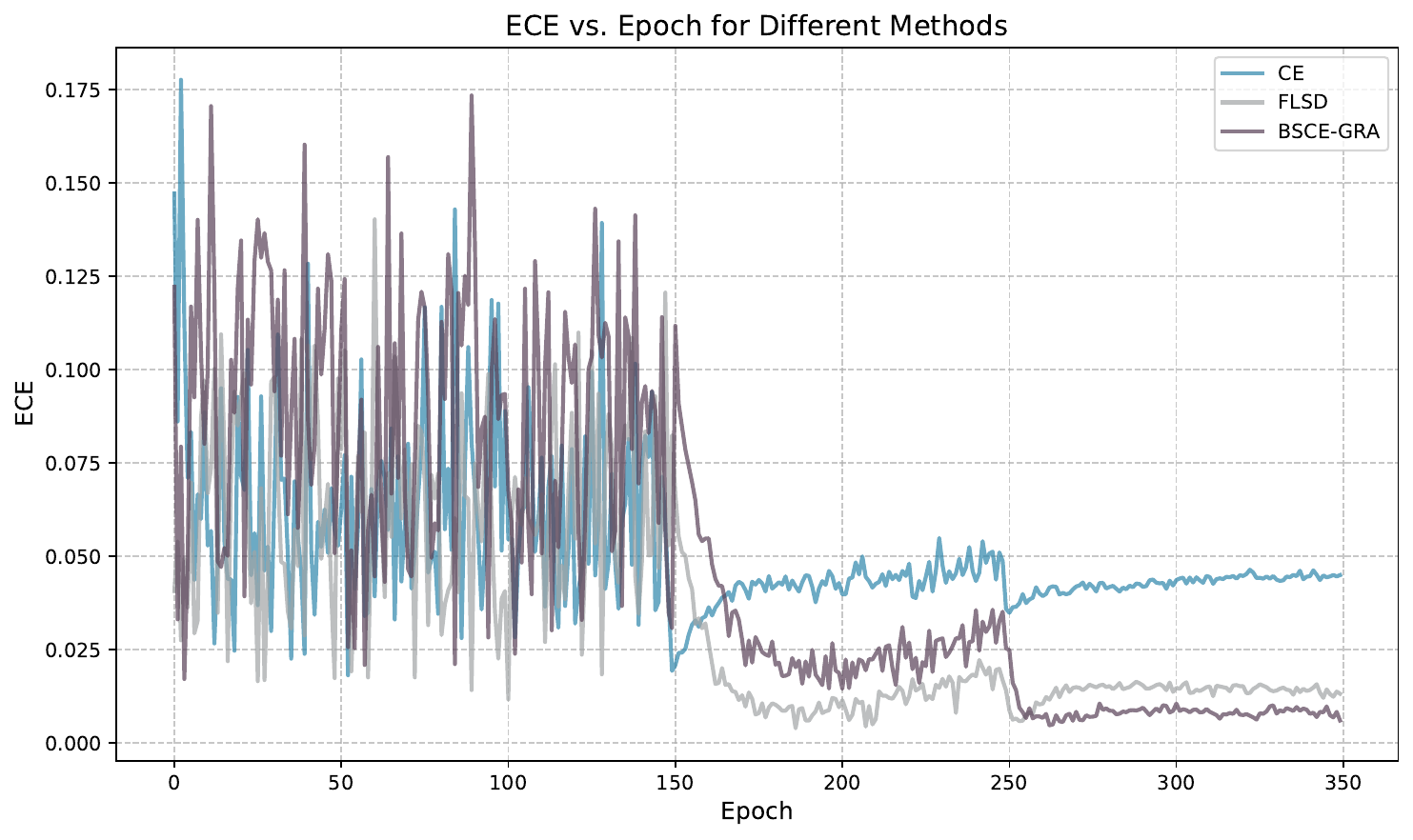}
        \caption{ECE over epochs using CE, Focal Loss and BSCE-GRA.}
        \label{fig: ece over epochs}
    \end{subfigure}
    \hfill
    \begin{subfigure}[b]{0.3\textwidth}
        \centering
        \includegraphics[width=\textwidth]{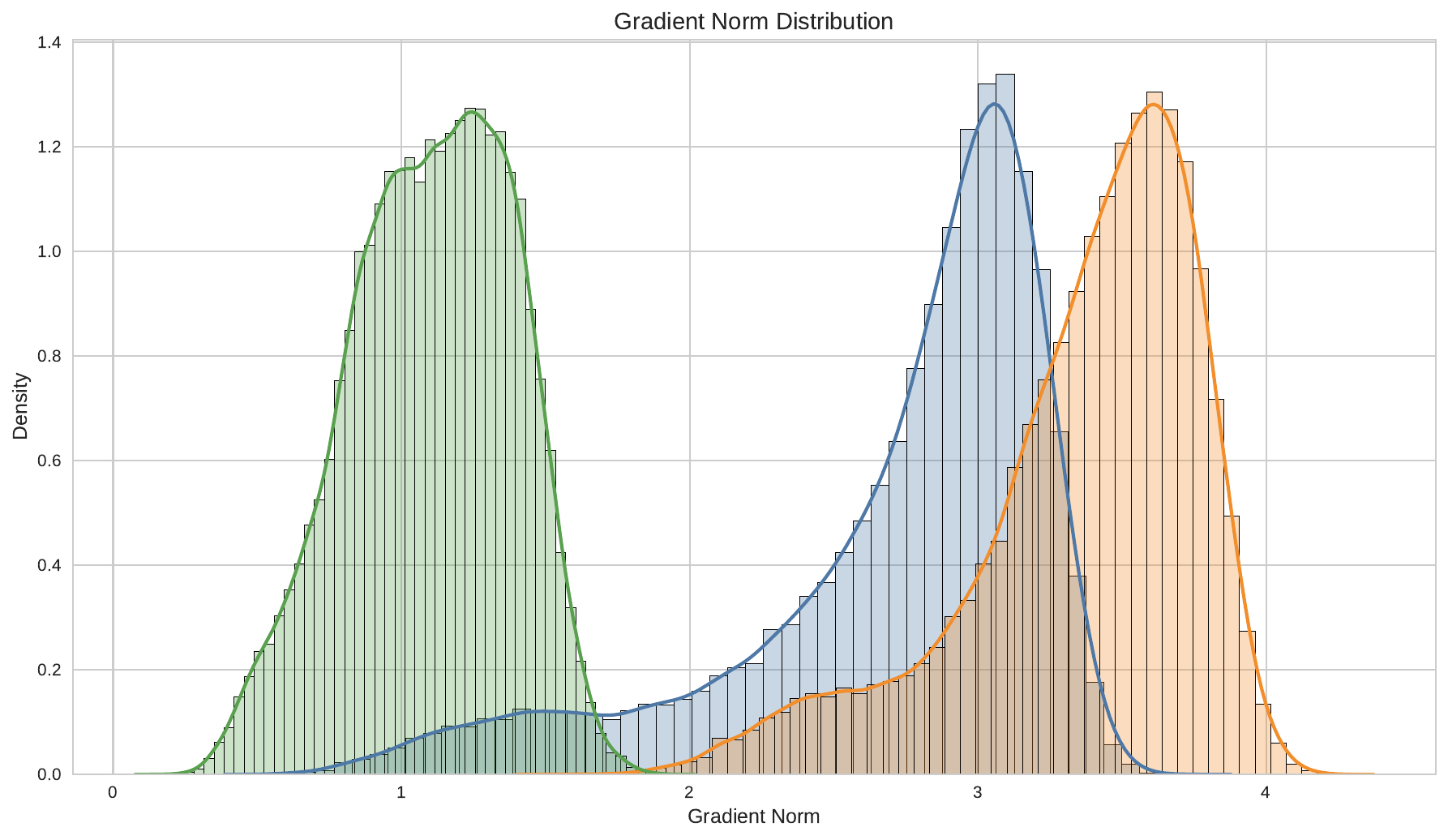}
        \caption{Gradient Density at epoch 150 using CE, Focal Loss and BSCE-GRA.}
        \label{fig: Gradient Density at epoch 150}
    \end{subfigure}
    \hfill
    \begin{subfigure}[b]{0.3\textwidth}
        \centering
        \includegraphics[width=\textwidth]{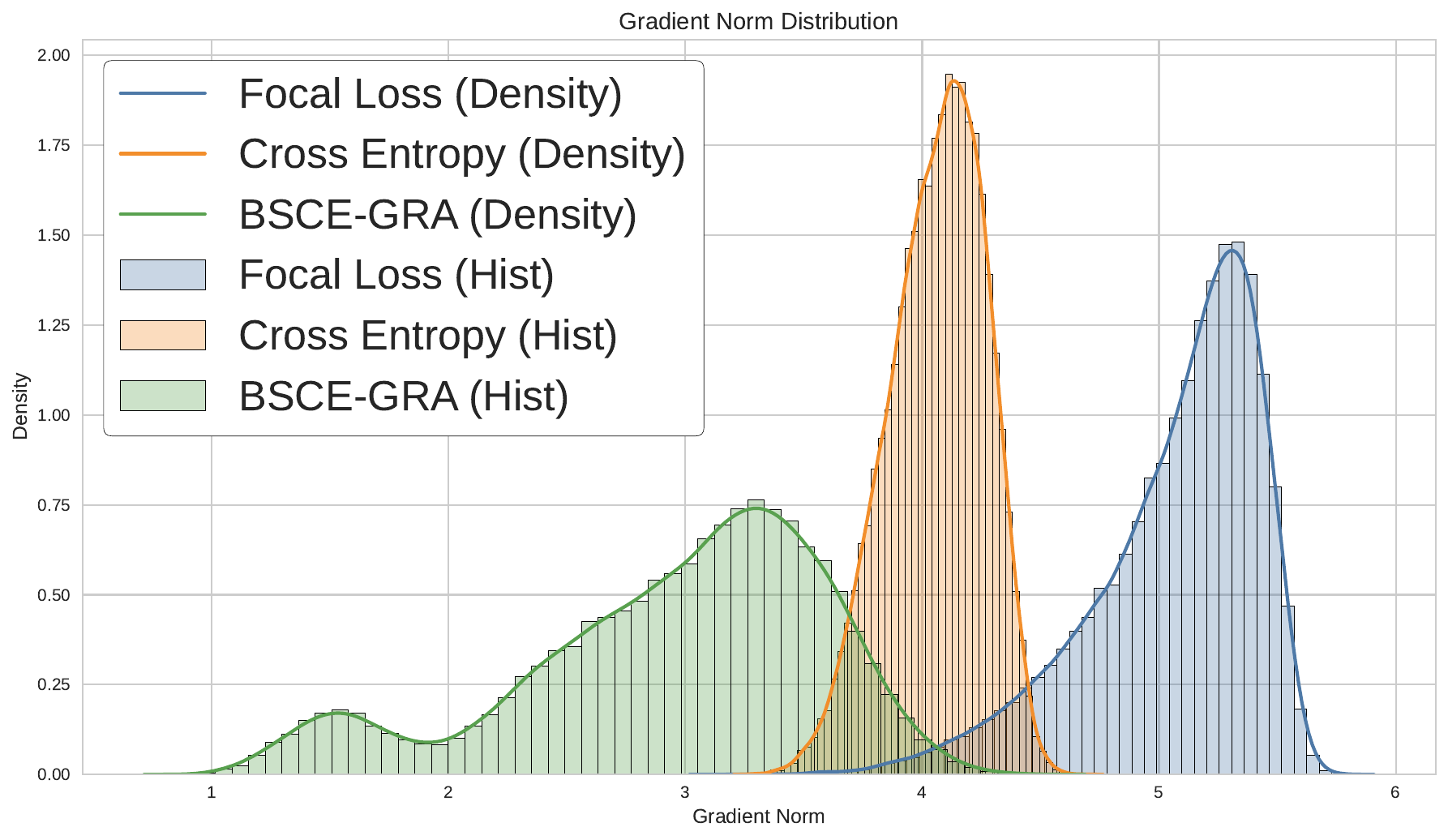}
        \caption{Gradient Density at epoch 250 using CE, Focal Loss and BSCE-GRA.}
        \label{fig: Gradient Density at epoch 250}
    \end{subfigure}
    \caption{Figure~\ref{fig: ece over epochs} presents the evolution of ECE throughout the training process, demonstrating that our method rapidly converges to the best result by epoch 250. The subsequent figures depict the gradient magnitudes of various methods between epochs 150 and 250.}
\end{figure*}

\noindent\textbf{Calibration over Training.}
Figure~\ref{fig: ece over epochs} presents the ECE on the test set for models trained with Focal Loss and Cross-Entropy loss over the entire training period on CIFAR-10 using ResNet-50.
To improve visualization, the ECE values are smoothed using an exponential moving average.
The figure suggests that, after the initial warm-up epochs, where predicted probabilities are unstable, the ECE of models trained with BSCE-GRA and Focal Loss consistently remains lower compared to models trained with Cross-Entropy.
It also indicates that, during training with a moderate learning rate from epochs 150 to 200, Focal Loss tends to produce better-calibrated models than BSCE-GRA.
This may be due to the fact that, during these epochs, the model makes more predictions with mid-range confidence levels, and as shown in Figure~\ref{fig: grad of focal loss vs confidence}, Focal Loss directs the model's attention towards these moderately uncertain samples.
In contrast, BSCE-GRA imposes stronger regularization on the gradient, resulting in smaller optimization steps compared to Focal Loss.
To further validate this hypothesis, we present the gradient density of the last linear layer across the entire training set at epochs 150 and 250 in Figure~\ref{fig: Gradient Density at epoch 150} and Fig.~\ref{fig: Gradient Density at epoch 250}, which shows that BSCE-GRA results in smaller gradient magnitudes during these epochs.
After 250 epochs, when the learning rate undergoes its second reduction, the model trained with BSCE-GRA soon achieves better calibration performance compared to FL and CE.

\subsection{Gradient Value among Uncertainty}
We extend our analysis to examine the relationship between gradient magnitude and uncertainty.
Specifically, we compute the gradient norms of the last linear layer for all samples in the training set at epochs 50, 150, 250, and 350.
To quantify uncertainty, we calculate the Brier Score for each training sample.
We present the results for Focal Loss (FL), Dual Focal Loss (DFL), and BSCE-GRA.
Figure~\ref{fig: gradient bs} illustrates the relationship between gradient magnitude and Brier Score.
The figure clearly shows that the gradients produced by BSCE-GRA are the most sensitive to changes in the Brier Score, as indicated by the narrow distribution compared to other loss functions.
This aligns with our goal: we aim for BSCE-GRA to exhibit sufficient sensitivity to uncertainty, allowing the model to adjust its gradient values based on changes in uncertainty, thereby focusing more on highly uncertain samples.
Notably, the gradient distribution for Dual Focal Loss is relatively more dispersed concerning the Brier Score, with a wider range of possible values compared to other loss functions. 
This could be attributed to the partial derivatives involving other classes, as Dual Focal Loss takes the second most probable class into account in its loss calculation.
In contrast, Focal Loss and BSCE-GRA only involve gradients along a single dimension.

\subsection{Weighting FL and DFL on Gradients}
We evaluate the performance of directly applying the uncertainty terms $u_{\text{\tiny FL}}$ and $u_{\text{\tiny DFL}}$ as weights on the gradients.
Experiments are conducted using the default settings discussed above, on CIFAR-10 with ResNet-50.
The results are presented in Table~\ref{tab: gradient or not experiment}.
It is evident that applying weights on gradients results in performance improvement for both $u_{\text{\tiny FL}}$ and $u_{\text{\tiny DFL}}$, further validating its effectiveness.

We conduct additional experiments to comprehensively validate our proposed method, BSCE-GRA, under various settings.
Due to page limitations, the extended experiments are provided in the Appendix.
\section{Conclusion}
In this paper, we proposed a novel approach to model calibration by directly weighting gradients based on uncertainty.
We analyzed the strengths and limitations of Focal Loss and Dual Focal Loss from the perspective of sample weighting and introduced a framework that scales gradient magnitudes based on model uncertainty to focus more on uncertain samples.
Additionally, we introduced BSCE-GRA, a loss function incorporating uncertainty metrics to enhance model calibration.
Extensive experiments on various datasets and network architectures demonstrated significant improvements in calibration performance, achieving state-of-the-art results. Our findings emphasize the value of integrating uncertainty-aware mechanisms directly into the optimization process, providing a reliable framework for training calibrated deep neural networks suitable for real-world applications requiring trustworthy predictions.
Future work may explore further optimization strategies for uncertainty-weighted approaches and their impact on tasks like active learning and robustness to adversarial attacks.

\section{Acknowledgment}
This work was supported in part by the Start-up Grant (No. 9610680) of the City University of Hong Kong, Young Scientist Fund (No. 62406265) of NSFC, and the Australian Research Council under Projects DP240101848 and FT230100549.
\clearpage
{
    \small
    \bibliographystyle{ieeenat_fullname}
    \bibliography{main}
}

\clearpage
\setcounter{page}{1}
\maketitlesupplementary

\begin{table*}
  \centering
  \resizebox{\textwidth}{!}{%
  \begin{tabular}{l c c c c c c c c}
    \toprule
    Dataset & Model & CE & BL & MMCE & FLSD & DFL & BSCE & BSCE-GRA \\
    \midrule
    \multirow{4}{*}{CIFAR10}
    & ResNet50   & 95.08 & 94.34 & 95.04 & 95.04 & 94.63 & 95.03 & 94.69 \\
    & ResNet110  & 94.84 & 94.41 & 94.91 & 94.76 & 94.79 & 94.88 & 94.72 \\
    & WideResNet & 96.03 & 95.88 & 95.74 & 95.75 & 95.82 & 95.78 & 95.77 \\
    & DenseNet   & 94.95 & 94.35 & 94.73 & 94.92 & 94.58 & 94.76 & 94.84 \\
    \midrule
    \multirow{4}{*}{CIFAR100}
    & ResNet50   & 77.22 & 72.47 & 77.49 & 77.69 & 76.70 & 77.12 & 76.84 \\
    & ResNet110  & 77.44 & 74.42 & 77.42 & 77.77 & 77.27 & 77.30 & 77.16 \\
    & WideResNet & 79.51 & 78.72 & 79.14 & 80.44 & 80.35 & 79.96 & 80.28 \\
    & DenseNet   & 76.76 & 73.32 & 76.07 & 77.29 & 77.02 & 76.82 & 76.96 \\
    \midrule
    \multirow{1}{*}{TinyImageNet}
    & ResNet50   & 49.88 & 27.66 & 48.81 & 51.98 & 51.04 & 50.06 & 50.21\\
    \bottomrule
  \end{tabular}%
  }
  \caption{\textbf{Comparison of Calibration Methods Using Accuracy Across Various Datasets and Models.}}
  \label{tab: accuracy results of experiment}
\end{table*}
\begin{table*}
  \centering
  \resizebox{\textwidth}{!}{%
  \begin{tabular}{l c c c c c c c c c c c c c c c c c}
    \toprule
    \multirow{2}{*}{Dataset} & \multirow{2}{*}{Model} 
    & \multicolumn{2}{c}{CE} & \multicolumn{2}{c}{BL} & \multicolumn{2}{c}{MMCE} & \multicolumn{2}{c}{FLSD} & \multicolumn{2}{c}{DFL} & \multicolumn{2}{c}{BSCE} & \multicolumn{2}{c}{BSCE-GRA}\\
    & & Pre T & Post T & Pre T & Post T & Pre T & Post T & Pre T & Post T & Pre T & Post T & Pre T & Post T & Pre T & Post T \\
    \midrule
    \multirow{4}{*}{CIFAR10}
    & ResNet50   & 4.34 & 2.09 & 4.28 & 1.87 & 4.47 & 2.11 & 1.56 & 1.45 & 1.22 & 1.22 & 0.96 & 0.96 & \textbf{0.71} & \textbf{0.71} \\
    & ResNet110  & 4.70 & 2.46 & 4.48 & 2.11 & 4.80 & 2.24 & 1.89 & 1.56 & \textbf{1.20} & \textbf{1.20 }& 1.38 & 1.38 & 1.28 & 1.28 \\
    & WideResNet & 3.35 & 1.87 & 2.86 & 1.82 & 3.62 & 1.98 & 1.92 & 1.57 & 3.12 & \textbf{1.43} & \textbf{1.72} & 1.53 & 1.76 & 1.60 \\
    & DenseNet   & 4.61 & 2.43 & 3.96 & 1.67 & 4.81 & 2.38 & 1.44 & 1.52 & \textbf{0.85} & \textbf{0.96} & 0.99 & 0.99 & 1.09 & 1.09 \\
    \midrule
    \multirow{4}{*}{CIFAR100}
    & ResNet50   & 18.04 & 3.84 & 7.86 & 4.27 & 15.85 & 3.28 & 5.50 & 2.76 & 2.68 & 2.85 & 2.22 & 2.22 & \textbf{1.82} & \textbf{1.82} \\
    & ResNet110  & 18.84 & 5.90 & 16.77 & 4.41 & 18.65 & 4.69 & 6.85 & 3.71 & 3.90 & 3.90 & 2.71 & 2.71 & \textbf{2.43} & \textbf{2.43} \\
    & WideResNet & 14.79 & 3.43 & 7.55 & 4.52 & 14.57 & 3.22 & 2.67 & 2.64 & 5.50 & 2.58 & 2.64 & \textbf{2.37 }& \textbf{2.48} & 2.48 \\
    & DenseNet   & 19.09 & 3.93 & 8.05 & 3.09 & 17.55 & 2.85 & 3.29 & \textbf{1.50} & 4.69 & 1.76 & 1.62 & 1.62 & \textbf{1.52} & 1.52 \\
    \midrule
    \multirow{1}{*}{TinyImageNet}
    & ResNet50   & 14.93 & 5.15 & 6.80 & 1.38 & 13.50 & 4.92 & \textbf{1.90} & 1.90 & 6.71 & 2.20 & 4.00 & 1.70 & 4.56 & \textbf{1.34 }\\
    \bottomrule
  \end{tabular}%
  }
  \caption{\textbf{Comparison of Calibration Methods Using AdaECE Across Various Datasets and Models.} AdaECE values are reported using adaptive binning, with the best-performing method for each dataset-model combination highlighted in bold. Results are averaged over three runs with different random seeds.}
  \label{tab: AdaECE results of experiment}
\end{table*}

\begin{table*}
  \centering
  \resizebox{\textwidth}{!}{%
  \begin{tabular}{l c c c c c c c c c c c c c c c c c}
    \toprule
    \multirow{2}{*}{Dataset} & \multirow{2}{*}{Model} 
    & \multicolumn{2}{c}{CE} & \multicolumn{2}{c}{BL} & \multicolumn{2}{c}{MMCE} & \multicolumn{2}{c}{FLSD} & \multicolumn{2}{c}{DFL} & \multicolumn{2}{c}{BSCE} & \multicolumn{2}{c}{BSCE-GRA}\\
    & & Pre T & Post T & Pre T & Post T & Pre T & Post T & Pre T & Post T & Pre T & Post T & Pre T & Post T & Pre T & Post T \\
    \midrule
    \multirow{4}{*}{CIFAR10}
    & ResNet50   & 0.90 & 0.46 & 0.90 & 0.50 & 0.92 & 0.51 & 0.41 & 0.42 & 0.40 & 0.40 & 0.38 & 0.38 & \textbf{0.38} & \textbf{0.38} \\
    & ResNet110  & 0.97 & 0.50 & 0.95 & 0.53 & 0.98 & 0.53 & 0.47 & 0.44 & 0.42 & 0.42 & 0.40 & 0.40 & \textbf{0.39} & \textbf{0.39} \\
    & WideResNet & 0.71 & 0.38 & 0.63 & 0.40 & 0.76 & 0.41 & 0.44 & 0.35 & 0.83 & 0.37 & 0.44 & 0.35 & \textbf{0.38} & \textbf{0.35} \\
    & DenseNet   & 0.96 & 0.52 & 0.85 & 0.49 & 0.99 & 0.53 & 0.41 & 0.38 & 0.42 & 0.38 & 0.40 & 0.40 & \textbf{0.37} & \textbf{0.37} \\
    \midrule
    \multirow{4}{*}{CIFAR100}
    & ResNet50   & 0.40 & 0.21 & 0.24 & 0.24 & 0.36 & 0.21 & 0.21 & 0.21 & 0.21 & 0.20 & 0.21 & 0.21 & \textbf{0.20} & \textbf{0.20} \\
    & ResNet110  & 0.41 & 0.22 & 0.38 & 0.23 & 0.41 & 0.21 & 0.22 & 0.22 & 0.21 & 0.21 & 0.21 & 0.21 & \textbf{0.21} & \textbf{0.21} \\
    & WideResNet & 0.33 & 0.21 & 0.22 & 0.22 & 0.32 & 0.21 & \textbf{0.18} & 0.19 & 0.23 & 0.19 & 0.20 & 0.19 & 0.19 & \textbf{0.19} \\
    & DenseNet   & 0.42 & 0.23 & 0.25 & 0.24 & 0.39 & 0.23 & \textbf{0.19} & 0.20 & 0.24 & 0.20 & 0.21 & 0.21 & 0.20 & \textbf{0.20} \\
    \midrule
    \multirow{1}{*}{TinyImageNet}
    & ResNet50   & 0.22 & 0.17 & 0.17 & \textbf{0.14} & 0.21 & 0.17 & \textbf{0.16} & 0.16 & 0.17 & 0.16 & 0.16 & 0.16 & 0.17 & 0.16 \\
    \bottomrule
  \end{tabular}%
  }
  \caption{\textbf{Comparison of Calibration Methods Using Classwise ECE Across Various Datasets and Models.} Classwise ECE values are reported for each dataset-model combination, with the best-performing method highlighted in bold. Results are averaged over three runs with different random seeds.}
  \label{tab: Classwise ECE results of experiment}
\end{table*}

\begin{figure*}
    \centering
    \begin{subfigure}[b]{0.45\textwidth}
        \centering
        \includegraphics[width=\textwidth]{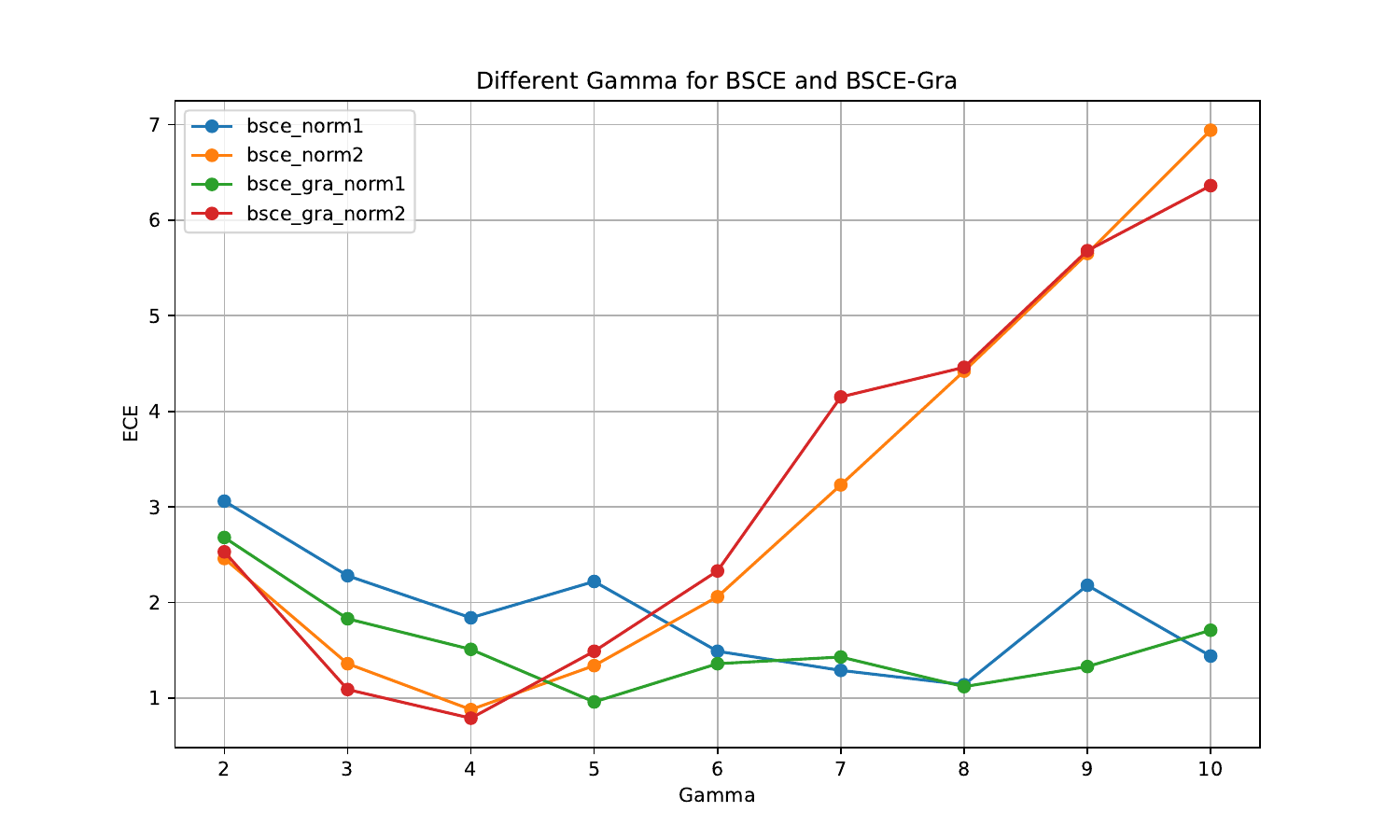}
        \caption{Different gamma and norm performs different bsce and bsce-gra ECE}
        \label{fig: bsce_gamma_ece}
    \end{subfigure}
    \hfill
    \begin{subfigure}[b]{0.45\textwidth}
        \centering
        \includegraphics[width=\textwidth]{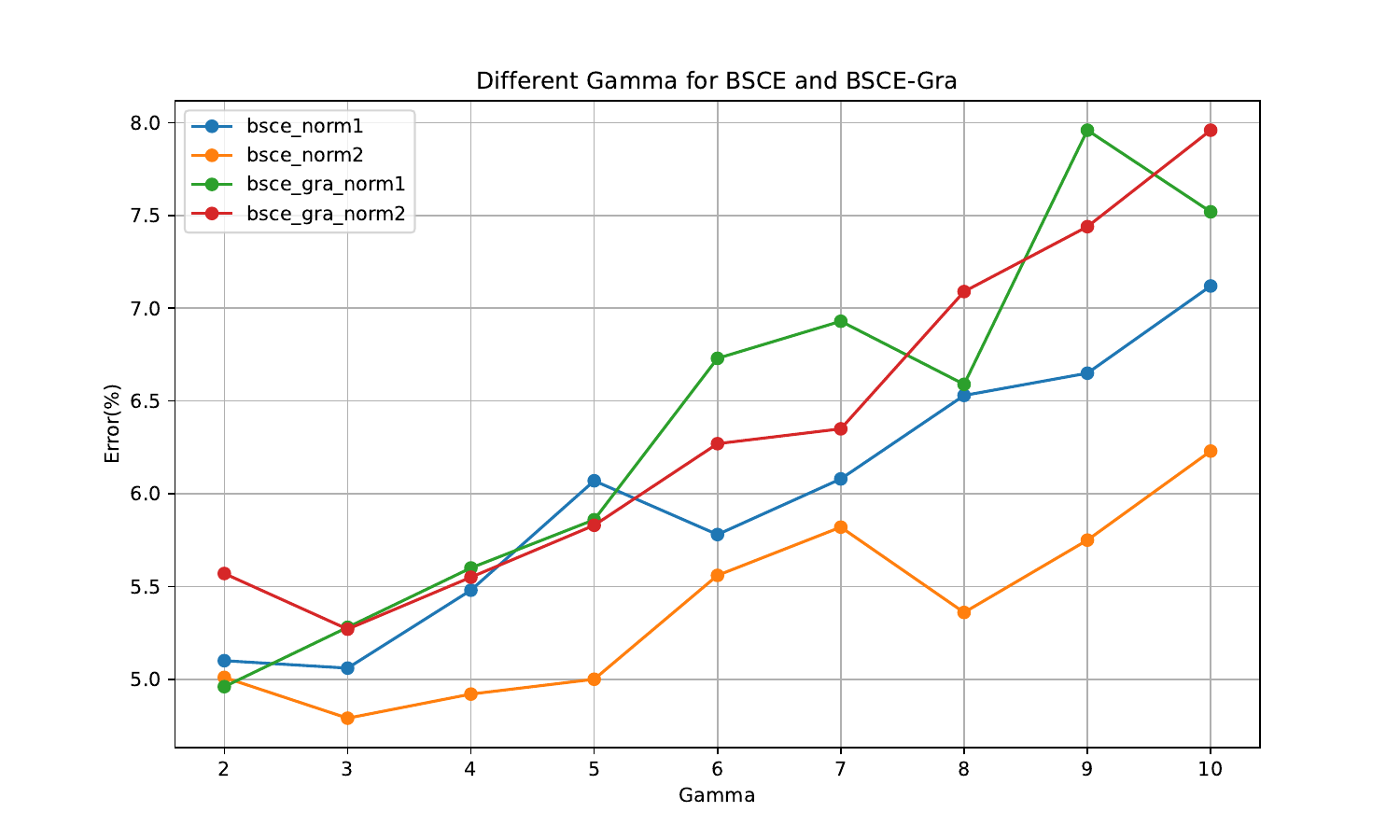}
        \caption{Different gamma and norm performs different bsce and bsce-gra Error}
        \label{fig: bsce_gamma_error}
    \end{subfigure}
\end{figure*}

\section{Proof of Equation~\ref{eq: effectiveness of BS}}
For the MSE term in Eq.~\ref{eq: effectiveness of BS}, we will have:
\begin{align}
    ||\hat{p}-y||^2=\sum_{k=1}^K (p_k^2-2p_ky_k+y_k),
\end{align}
as the $y$ is a one-hot vector.
Besides, the one-hot labels $y$ are sampled from Bernoulli Distributions: $y_k\sim\text{Bernoulli}(\eta_k(x))$, and the expectation of MSE can be computed as: 
\begin{align}
    \mathbb{E}_{y\sim\eta(x)}[||\hat{p}-y||^2]=\sum^K_{k=1}(p_k^2-2p_k\mathbb{E}[y_k]+\mathbb{E}[y_k]).
\end{align}
Since $\mathbb{E}[y_k]=\eta_k$, we will have:
\begin{align}
    c(\vx)-u_{\text{\tiny BS}}(\hat{p}(\vx))&=\mathbb{E}_{y\sim\eta(x)}[\Vert\hat{p}(\vx)-\rveta(\vx)\Vert^2_2-\Vert\hat{p}(\vx)-y\Vert^2_2]\notag \\ 
                       &=\sum^K_{k=1}\rveta_k(\vx)(\rveta_k(\vx)-1).
\end{align}

\section{Theoretical Evidence for the Effectiveness of BSCE-GRA}

Here, we prove that under strict convergence, the $K$-class predicted probability $q$ equals the actual class-posterior probability $\eta$, thereby preventing over/under-confidence.
For BSCE-GRA, we introduce the Lagrangian equation of BSCE-GRA as
\begin{align}
    L=[\sum_{i=1}^K (q_i-\eta_i)^2](-\sum_{i=1}^K\eta_i\log q_i)+\mu (\sum_{i=1}^K q_i-1)
\end{align} under the constraint $\sum_{i=1}^K q_i=1$, where $[\cdot]$ denotes detaching the gradient.
Since the MSE term can be considered as a constant $C$, considering the derivatives w.r.t $q_i^*$ as $0$, we have $\mu=C\eta_i/q_i$ for any $i$.
Therefore, for any class $i$, there exists a constant $k$ s.t. $q_i=k\eta_i$.
Considering the constraint $\sum_{i=1}^K q_i=1$, we find that $k=1$ and thus $q_i=\eta_i$, which implies that this is an optimal minimum solution.
When $q=\eta$, both MSE and CE equal $0$.
When $q\neq\eta$, BSCE-GRA$>0$.
Thus, the BSCE-GRA achieves a minimum when $q=\eta$.

We consider the extreme case for further evidence.
When $\eta_i=1$, $L(q)\supset(q_i-1)^2(-\log q_i)$.
The loss becomes $0$ when $q_i=\eta_i=1$.
When $\eta_i=0$, $L(q)\supset(q_i-0)^2(-0\cdot\log q_i)=0$.
For all classes, $\sum_{i=1}^K q_i=1$.
Therefore, $q_i=0$ is the optimal solution for the class where $\eta_i = 0$.
The optimal solution $q = \eta$ ensures the mitigation of over/under-confidence.

\section{Dataset Desciption}

We evaluate the performance of our proposed method, BSCE-GRA on multiple datasets to assess its calibration capabilities and robustness.
The datasets include CIFAR-10/100~\citep{krizhevsky2009learning} and Tiny-ImageNet~\citep{deng2009imagenet}.
Below, we provide specific details for each dataset used:

\textbf{CIFAR-10}
CIFAR-10 consists of 60,000 $32\times32$ color images divided into 10 classes, with 6,000 images per class (50,000 training and 10,000 test images).
The classes include airplanes, cars, birds, cats, deer, dogs, frogs, horses, ships, and trucks.
This dataset is widely used in image classification tasks due to its simplicity and balanced class distribution.
For our evaluation, we use 5,000 images from the training set for validation, ensuring a balanced split between training and validation data.

\textbf{CIFAR-100.}
CIFAR-100 follows a similar structure to CIFAR-10 but with 100 classes, each containing 600 images (500 training and 100 test images per class).
The classes in CIFAR-100 are more fine-grained compared to CIFAR-10, making it a more challenging dataset for image classification.
Each class belongs to one of 20 superclasses, adding an additional layer of complexity to the classification task. This dataset allows us to evaluate the performance of our methods on a more complex and diverse set of visual categories.

\textbf{Tiny ImageNet.}
Tiny ImageNet is a subset of the larger ImageNet dataset, consisting of 100,000 images across 200 classes, with each image resized to $64\times64$ pixels.
Each class contains 500 training images, 50 validation images, and 50 test images.
Tiny ImageNet is commonly used for benchmarking image classification models, providing a challenging task due to the increased number of classes compared to CIFAR-10/100 and the reduced image resolution compared to the original ImageNet dataset.
The diversity and scale of Tiny ImageNet make it suitable for evaluating the robustness and scalability of our proposed methods.

\section{Comparison Methods}
To assess the effectiveness of our proposed algorithm, we compare it against several established methods.
Details of these comparison methods are provided below:

\textbf{Brier Loss~\citep{brier1950verification}.}
Brier Loss calculates the squared error between the softmax logits and the one-hot encoded labels.
It serves as a measure of both model calibration and accuracy.

\textbf{MMCE Loss~\citep{kumar2018trainable}.}
Maximum Mean Calibration Error (MMCE) is a kernel-based auxiliary loss used alongside Negative Log-Likelihood (NLL) to enhance calibration performance.
It leverages a Reproducing Kernel Hilbert Space (RKHS) to evaluate and reduce miscalibration during training.

\textbf{Focal Loss~\citep{mukhoti2020calibrating}.}
FLSD-53 is a simplified version of the sample-dependent gamma ($\gamma$) approach in Focal Loss.
\citet{mukhoti2020calibrating} introduced a scheduling mechanism for gamma, replacing the original fixed value.
Specifically, they set $\gamma_{\text{focal}} = 5$ for $\hat{p}_c\in[0,0.2)$ and $\gamma_{\text{focal}} = 3$ for $\hat{p}_c\in[0.2,1]$.

\textbf{Dual Focal Loss~\citep{tao2023dual}.}
Dual Focal Loss (DFL) extends Focal Loss by incorporating the second highest predicted probability into the uncertainty metric.
This helps mitigate model underconfidence and improve calibration.
In our experiments, we set $\gamma_{\text{\tiny DualFocal}} = 5$ as suggested by their reported findings.

\section{Performance on Different Metrics}

We report the accuracy of each method in different settings in Table~\ref{tab: accuracy results of experiment}.
Although BSCE-GRA has the best calibration performance according to the ECE results in Table~\ref{tab: ECE results of experiment}, it shows a competitive performance in accuracy compared with other methods.
Adaptive-ECE is a calibration performance measure designed to address the bias inherent in the equal-width binning scheme used by ECE.
It adapts the bin size based on the number of samples, ensuring an even distribution of samples across bins.
The formula for Adaptive-ECE is as follows:
\begin{equation}
    \text{Adaptive-ECE}=\sum_{i=1}^B\frac{|B_i|}{N}|I_i-C_i|\ \text{s.t.} \forall i,j\cdot|B_i|=|B_j|
    \label{eq: adaptive ece}
\end{equation}
Table~\ref{tab: AdaECE results of experiment} shows that BSCE-GRA has the most optimal case compared to other methods, especially in the CIFAR100.
Classwise-ECE is an alternative measure of calibration performance that overcomes the limitation of ECE, which only evaluates the calibration of the predicted class. It can be formulated as:
\begin{equation}
    \text{Classwise-ECE}=\frac{1}{K}\sum_{i=1}^B\sum_{j=1}^K \frac{|B_{i,j}|}{N}|I_{i,j}-C_{i,j}|
\end{equation}
where $B_{i,j}$ denotes the set of samples with the $j^{\text{th}}$ class label in the $i^{\text{th}}$ bin, $I_{i,j}$ and $C_{i,j}$ represents the accuracy and confidence of samples in $B_{i,j}$.
Table~\ref{tab: Classwise ECE results of experiment} indicates that all methods perform similarly in terms of Classwise-ECE, yet BSCE-GRA consistently achieves the best results across most settings.

\section{Reliability Diagram Variants Across Different Settings}
\begin{figure*}[t]
    \centering
    % 第0行：Cross-Entropy
    \begin{subfigure}[b]{0.43\textwidth}
        \centering
        \includegraphics[width=\textwidth]{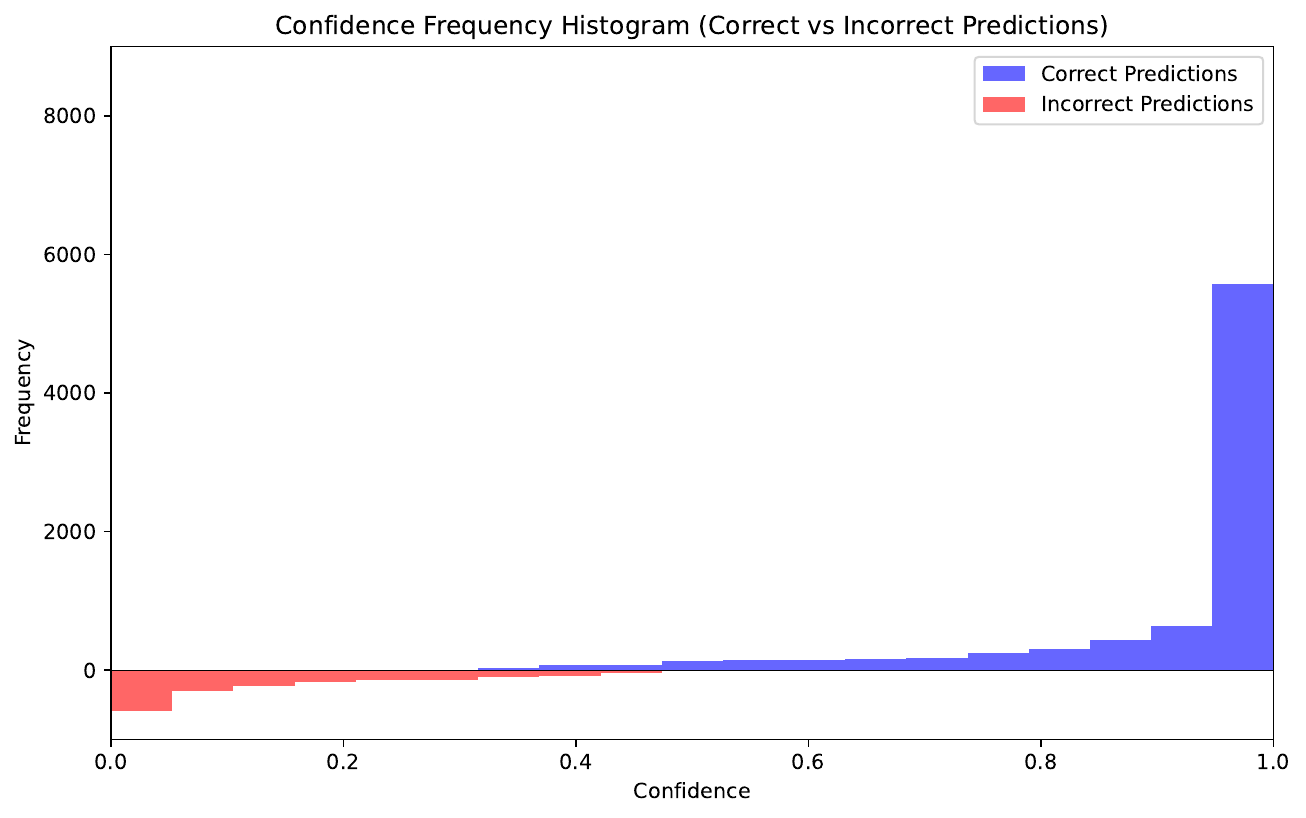}
        \caption{CE at epoch 50}
    \end{subfigure}
    \hfill
    \begin{subfigure}[b]{0.43\textwidth}
        \centering
        \includegraphics[width=\textwidth]{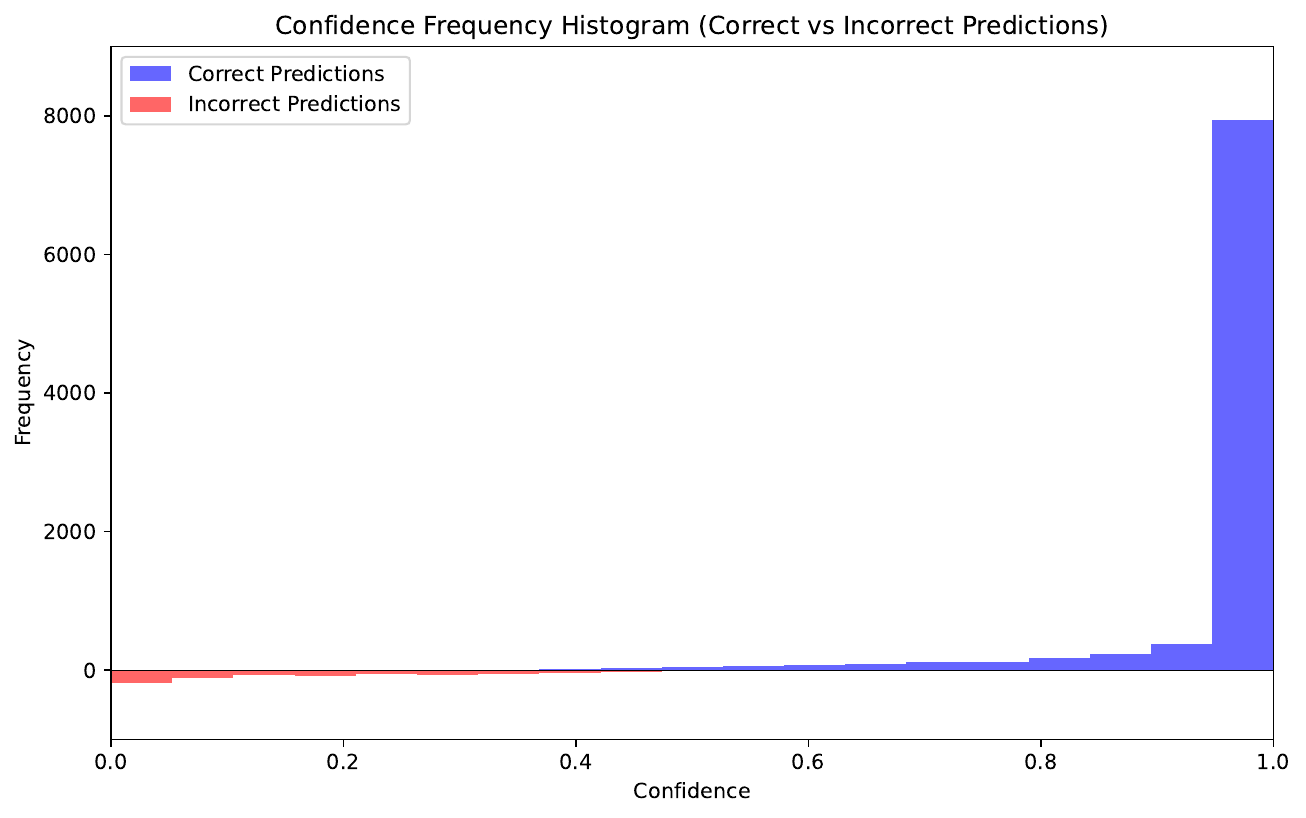}
        \caption{CE at epoch 150}
    \end{subfigure}

    \begin{subfigure}[b]{0.43\textwidth}
        \centering
        \includegraphics[width=\textwidth]{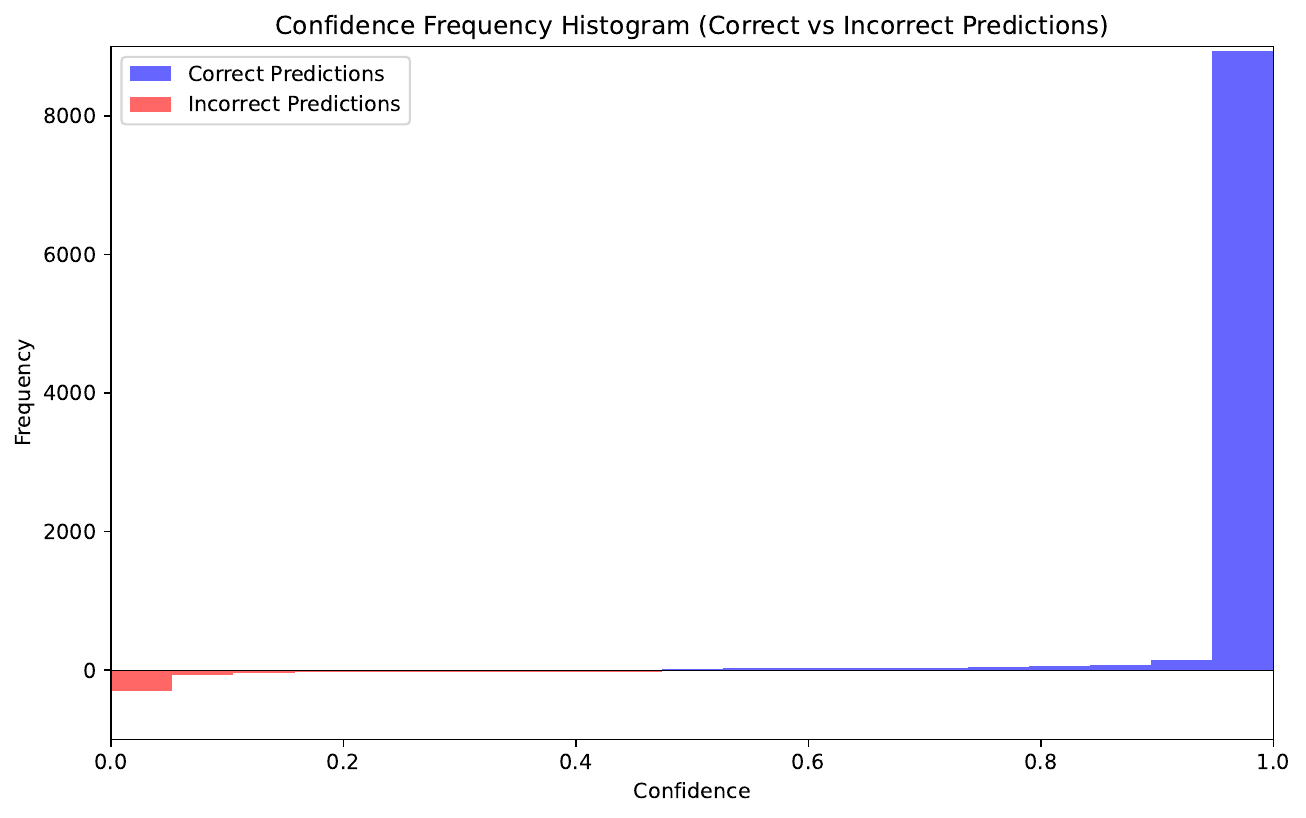}
        \caption{CE at epoch 250}
    \end{subfigure}
    \hfill
    \begin{subfigure}[b]{0.43\textwidth}
        \centering
        \includegraphics[width=\textwidth]{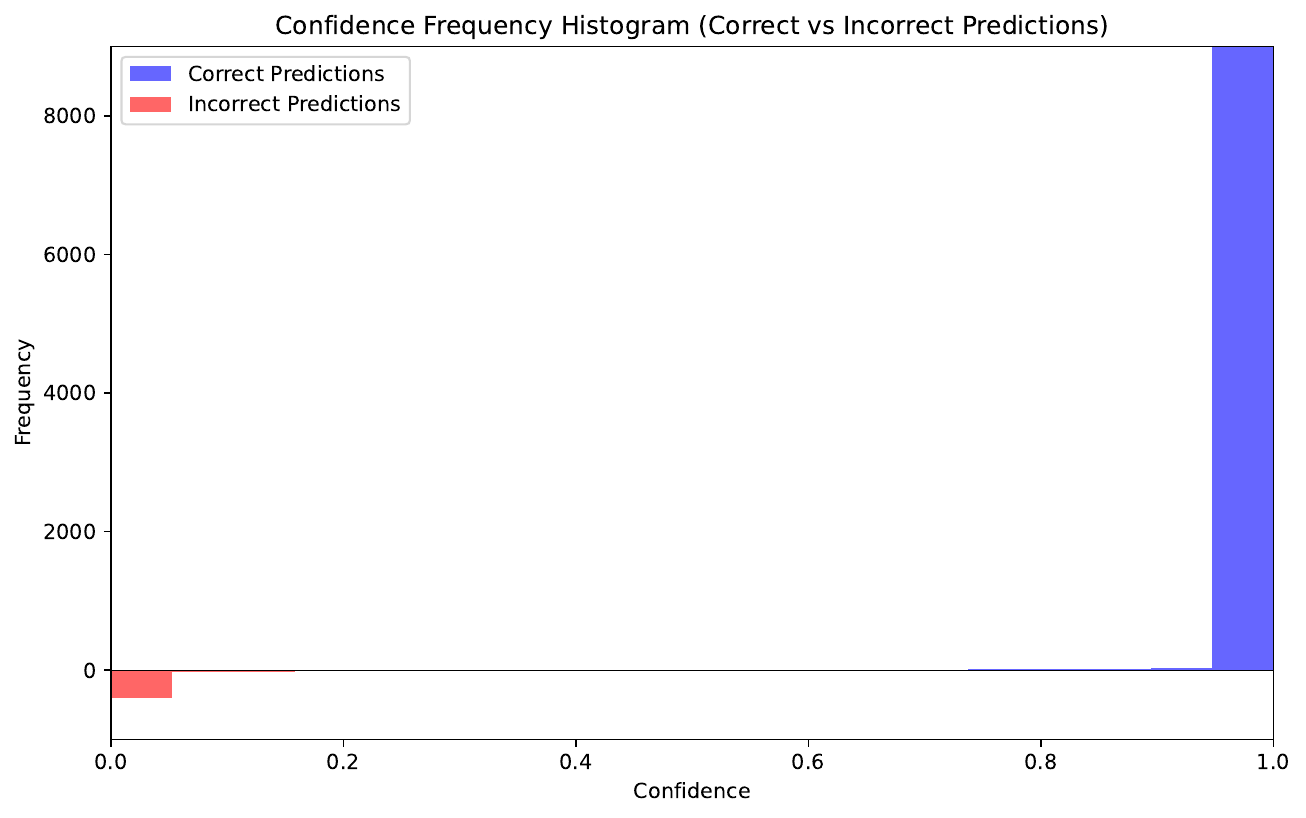}
        \caption{CE at epoch 350}
    \end{subfigure}
    
    % 第一行：FL
    \begin{subfigure}[b]{0.43\textwidth}
        \centering
        \includegraphics[width=\textwidth]{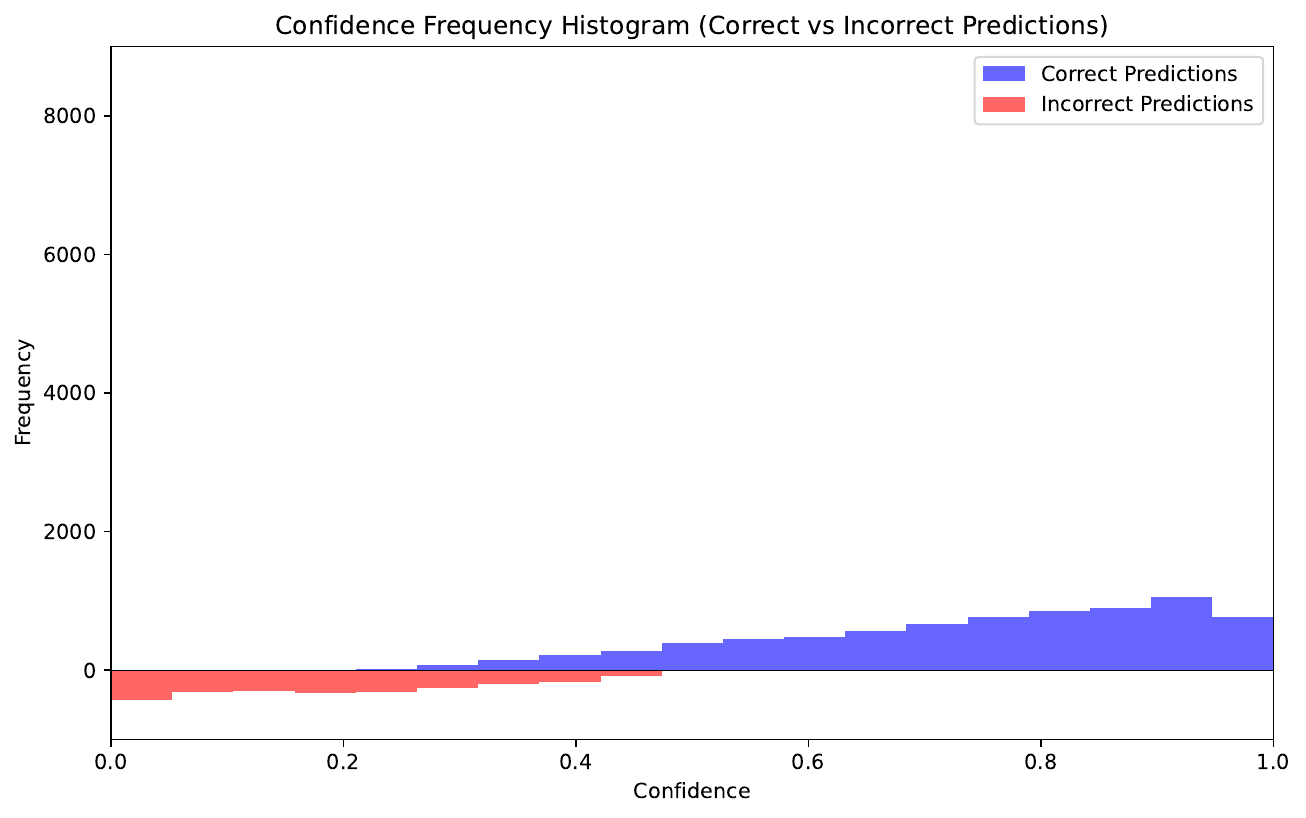}
        \caption{FL at epoch 50}
    \end{subfigure}
    \hfill
    \begin{subfigure}[b]{0.43\textwidth}
        \centering
        \includegraphics[width=\textwidth]{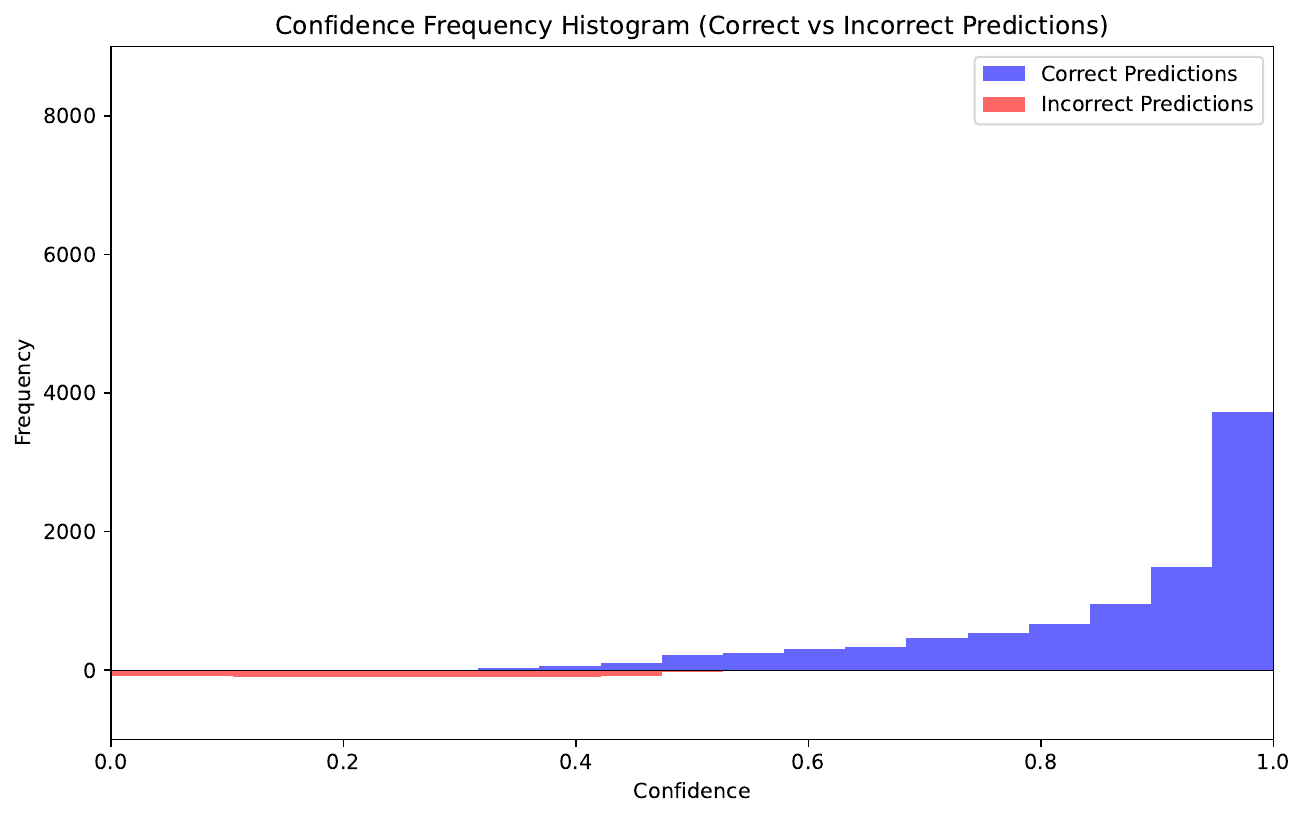}
        \caption{FL at epoch 150}
    \end{subfigure}

    \begin{subfigure}[b]{0.43\textwidth}
        \centering
        \includegraphics[width=\textwidth]{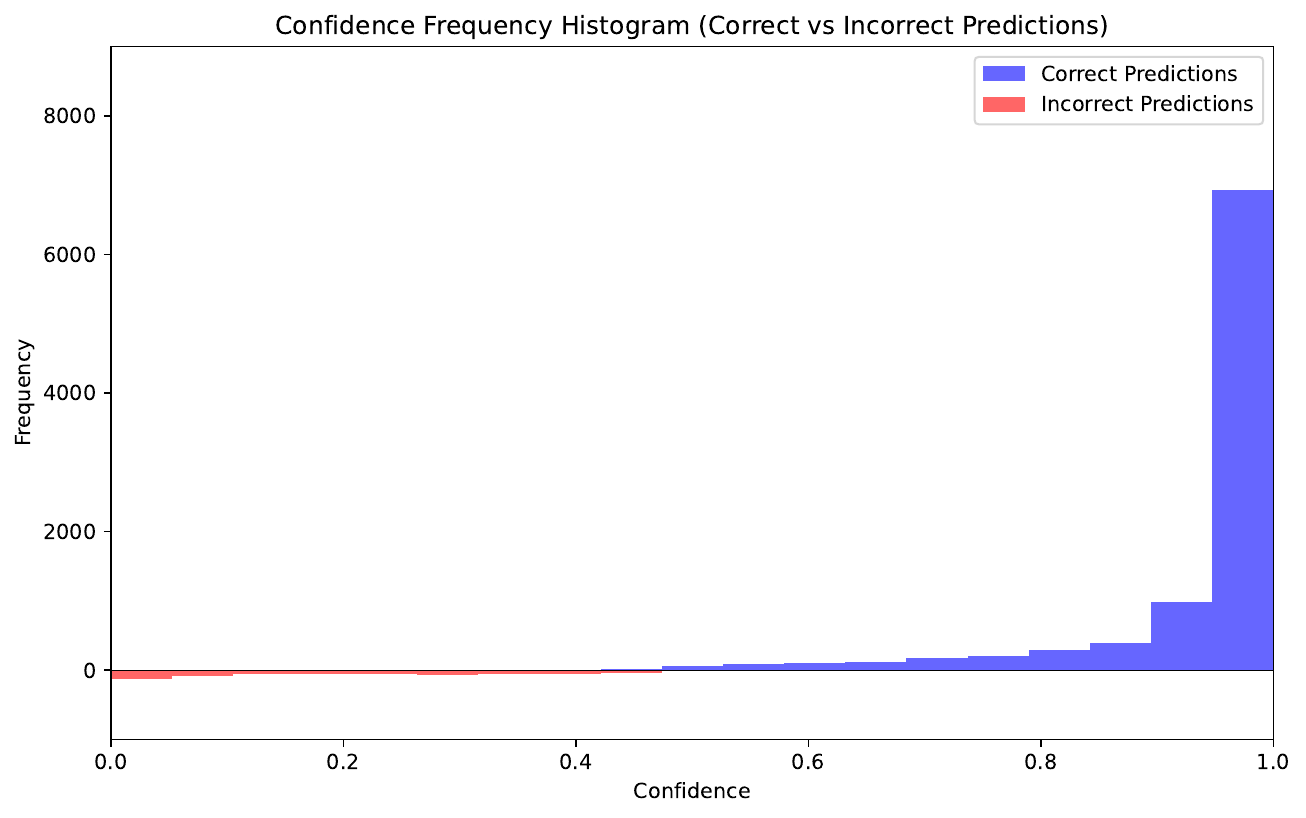}
        \caption{FL at epoch 250}
    \end{subfigure}
    \hfill
    \begin{subfigure}[b]{0.43\textwidth}
        \centering
        \includegraphics[width=\textwidth]{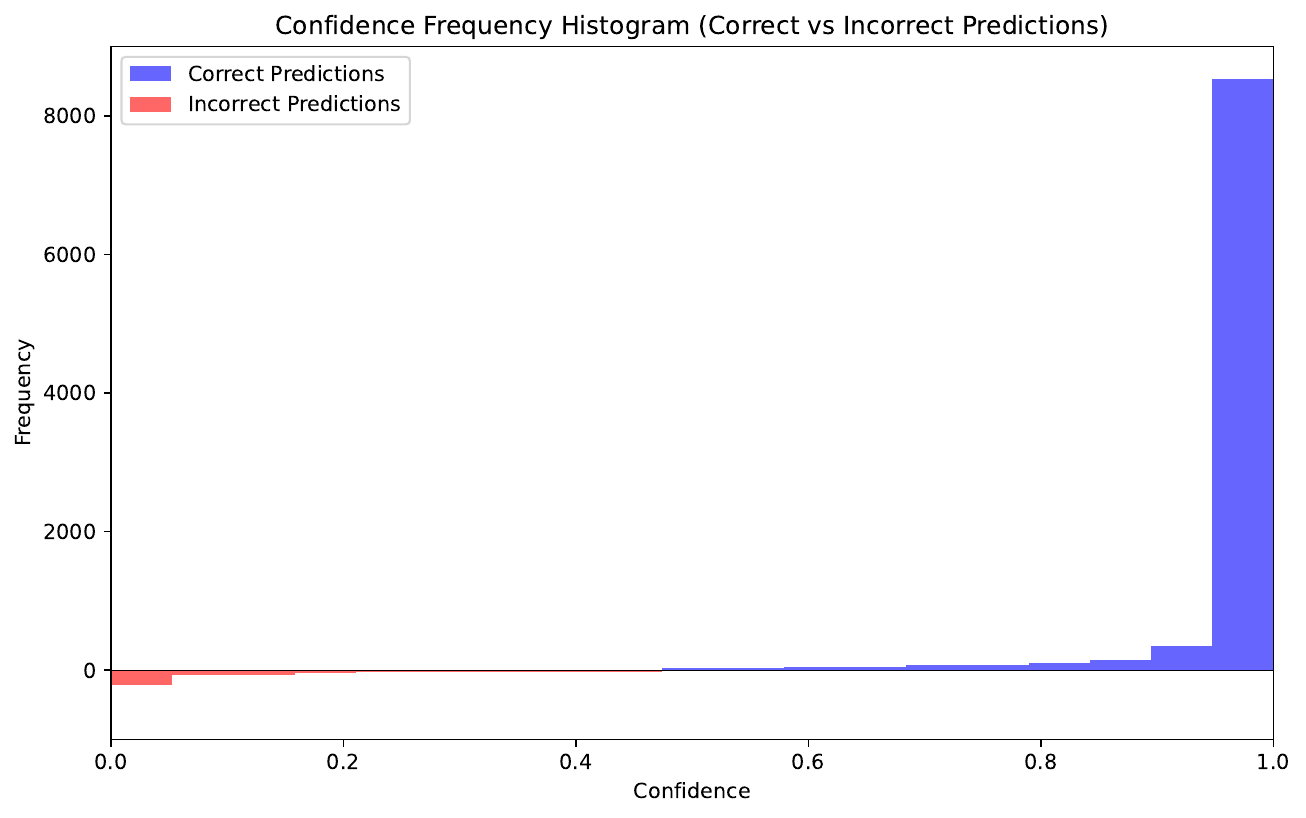}
        \caption{FL at epoch 350}
    \end{subfigure}
    \caption{Correct and Wrong Predictions with Cross Entropy and Focal Loss among different epochs.}
    \label{fig: confidence distribution1}
\end{figure*}

\begin{figure*}[t]
    \centering
    % 第二行：DFL
    \begin{subfigure}[b]{0.43\textwidth}
        \centering
        \includegraphics[width=\textwidth]{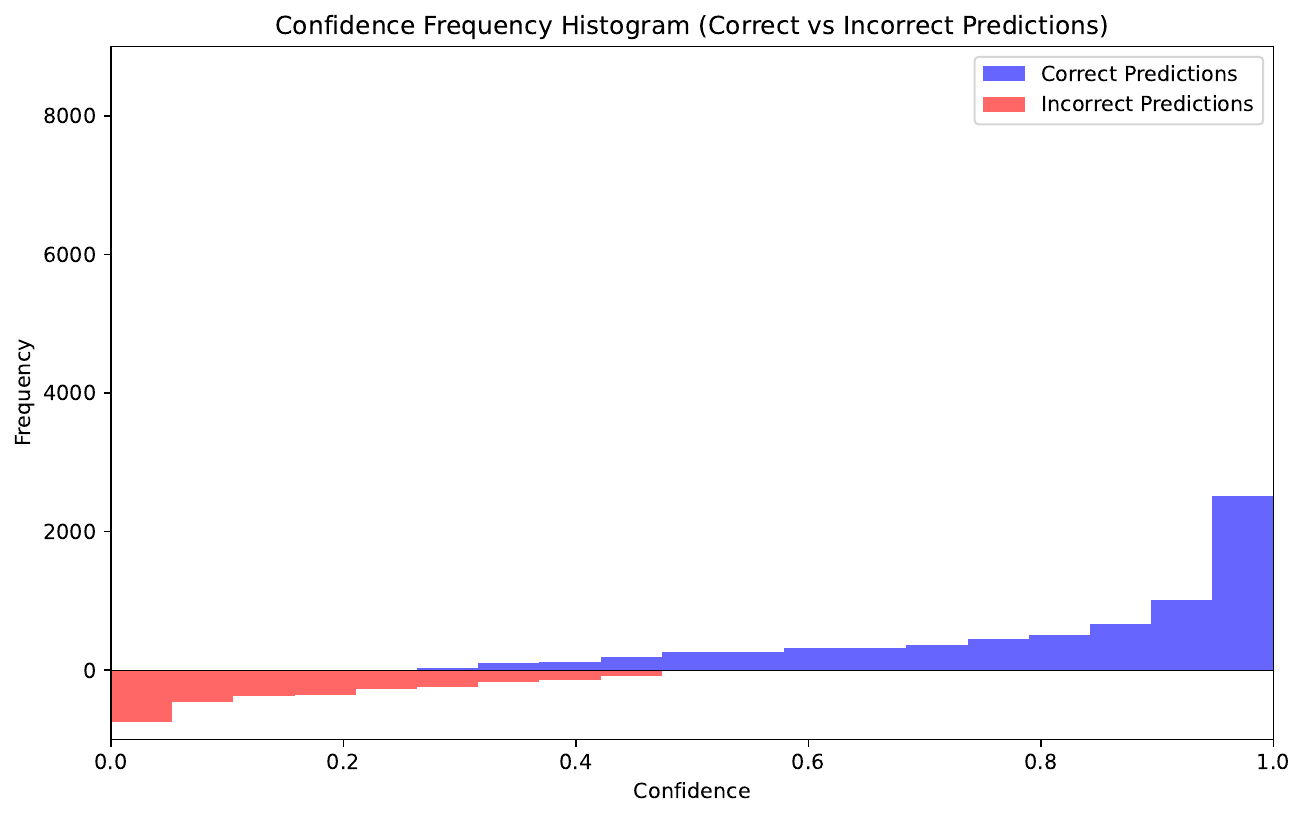}
        \caption{DFL at epoch 50}
    \end{subfigure}
    \hfill
    \begin{subfigure}[b]{0.43\textwidth}
        \centering
        \includegraphics[width=\textwidth]{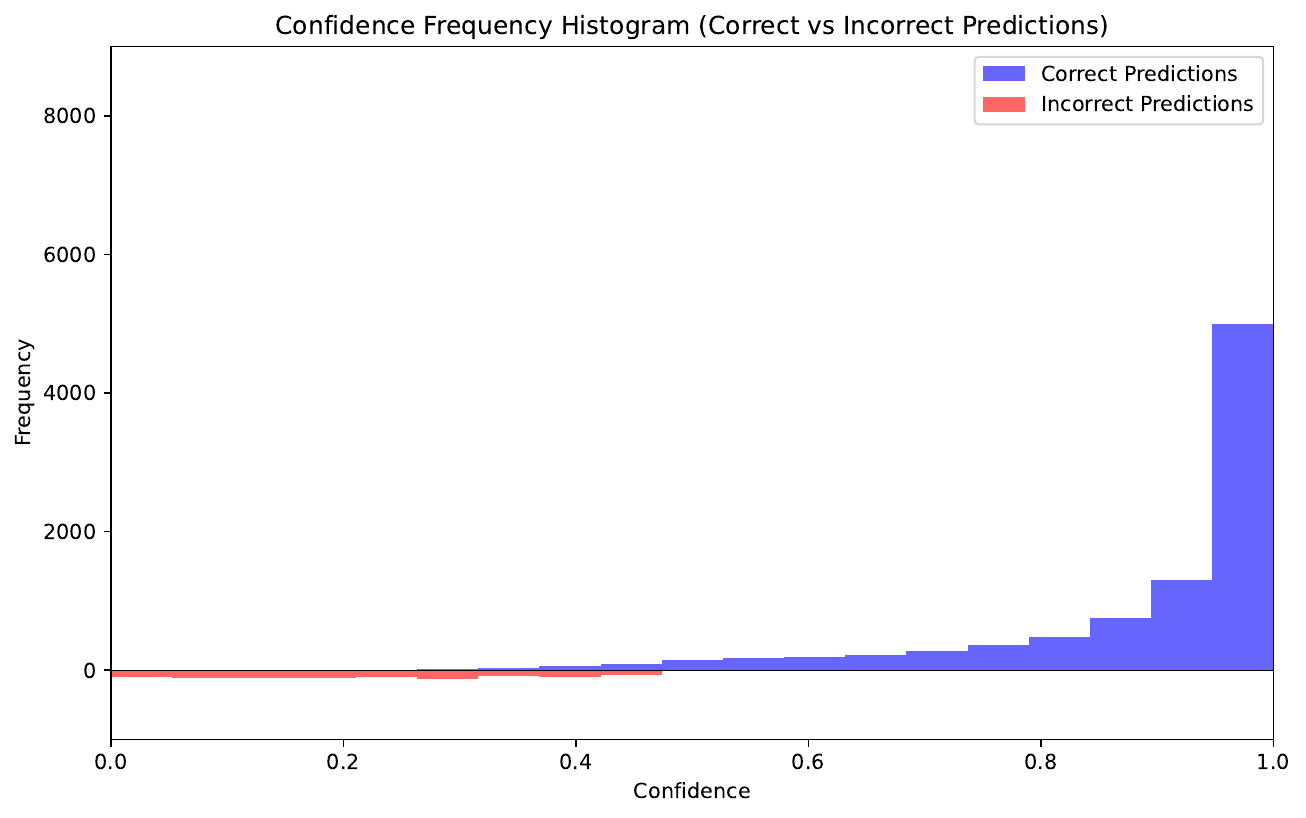}
        \caption{DFL at epoch 150}
    \end{subfigure}

    \begin{subfigure}[b]{0.43\textwidth}
        \centering
        \includegraphics[width=\textwidth]{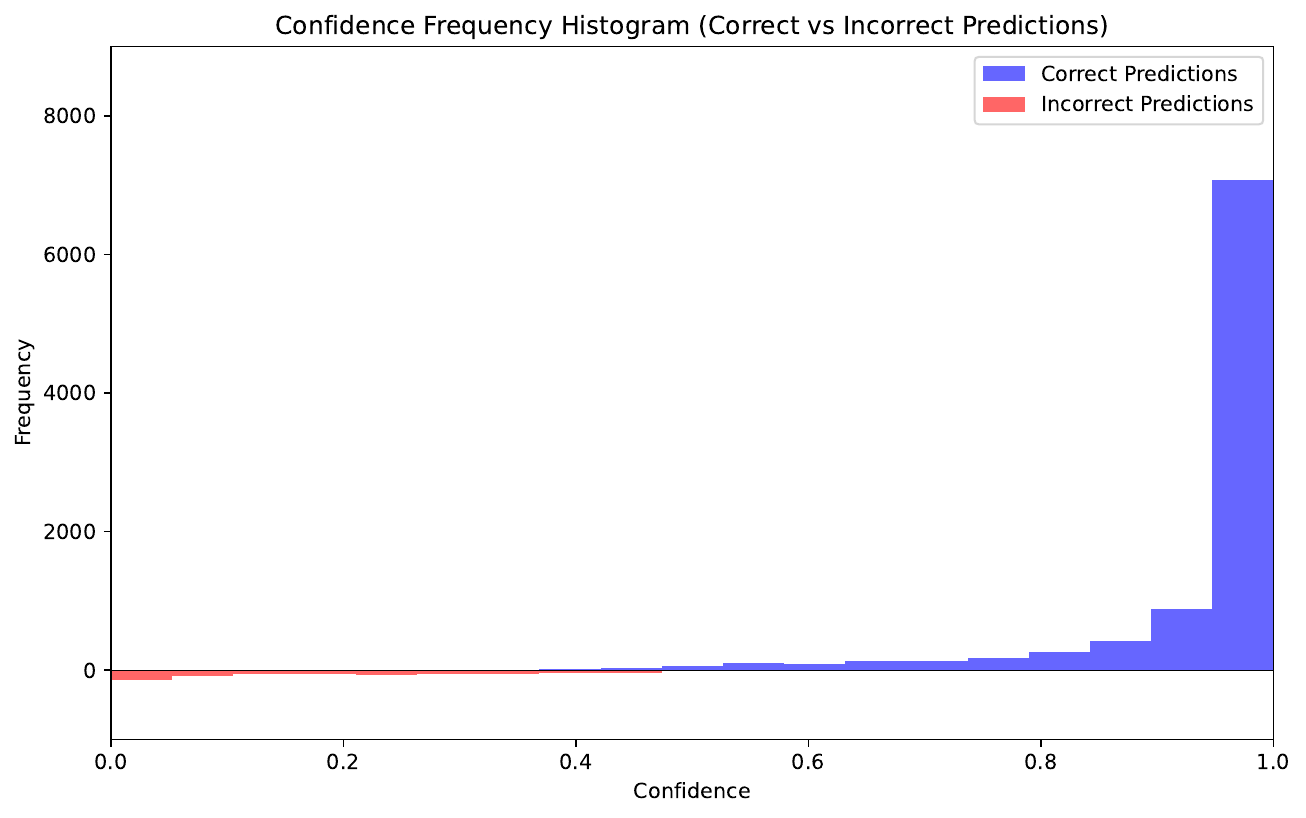}
        \caption{DFL at epoch 250}
    \end{subfigure}
    \hfill
    \begin{subfigure}[b]{0.43\textwidth}
        \centering
        \includegraphics[width=\textwidth]{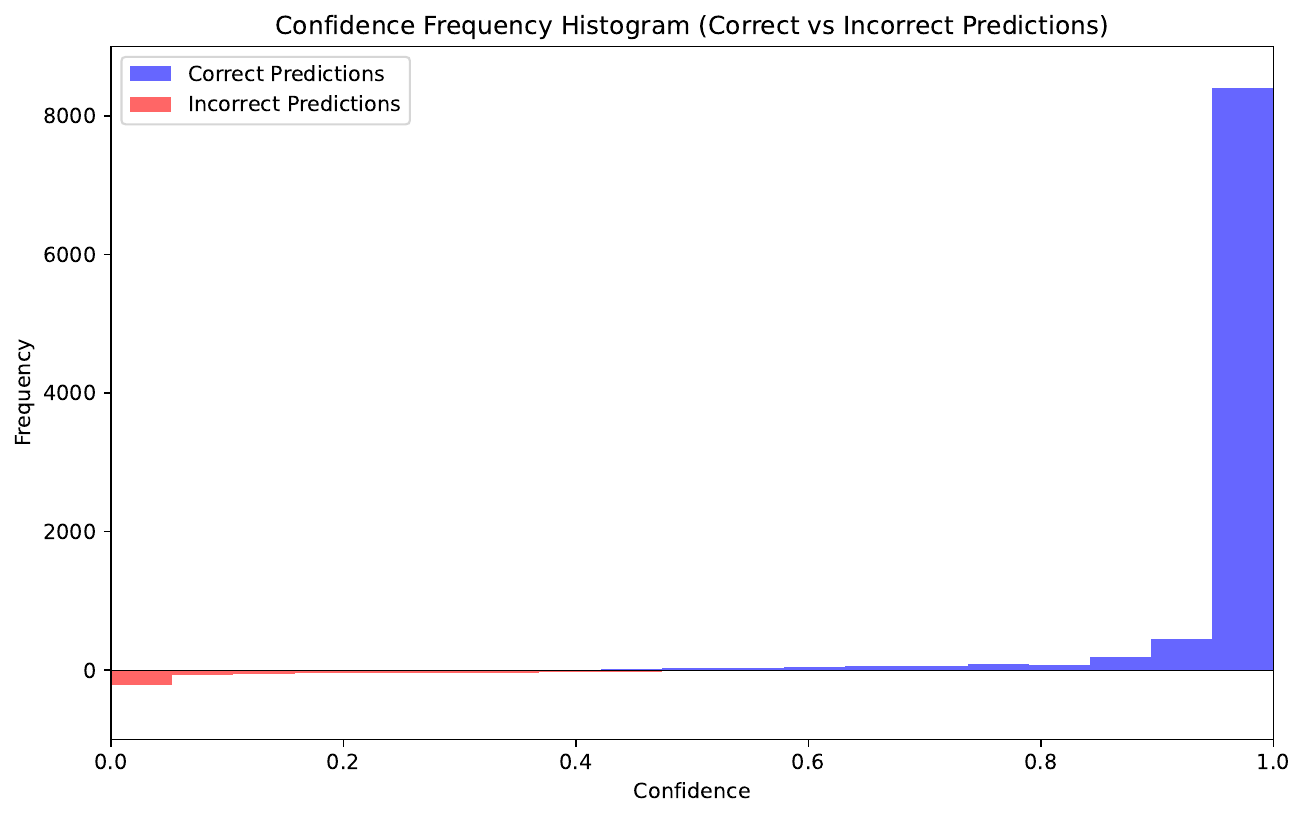}
        \caption{DFL at epoch 350}
    \end{subfigure}
    
    % 第三行：BSCE
    \begin{subfigure}[b]{0.43\textwidth}
        \centering
        \includegraphics[width=\textwidth]{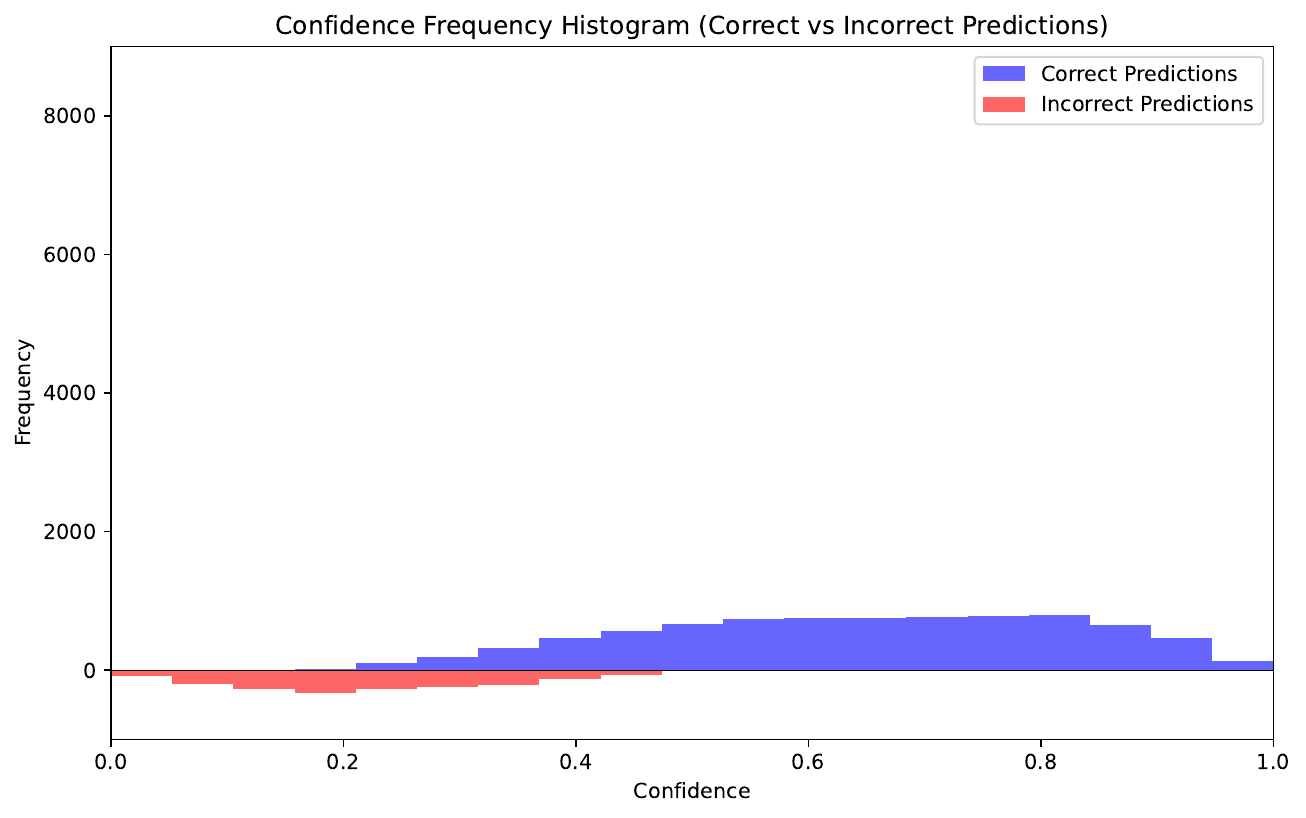}
        \caption{BSCE at epoch 50}
    \end{subfigure}
    \hfill
    \begin{subfigure}[b]{0.43\textwidth}
        \centering
        \includegraphics[width=\textwidth]{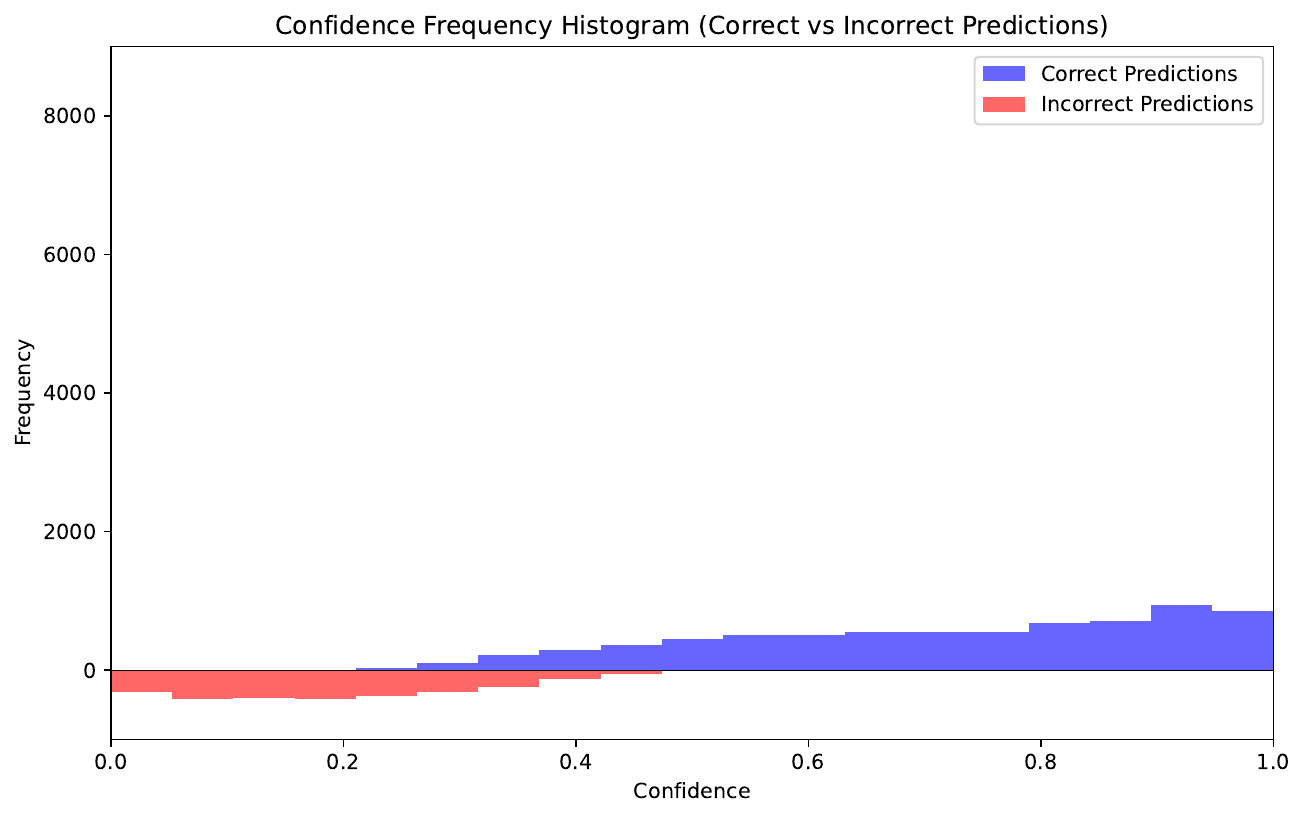}
        \caption{BSCE at epoch 150}
    \end{subfigure}

    \begin{subfigure}[b]{0.43\textwidth}
        \centering
        \includegraphics[width=\textwidth]{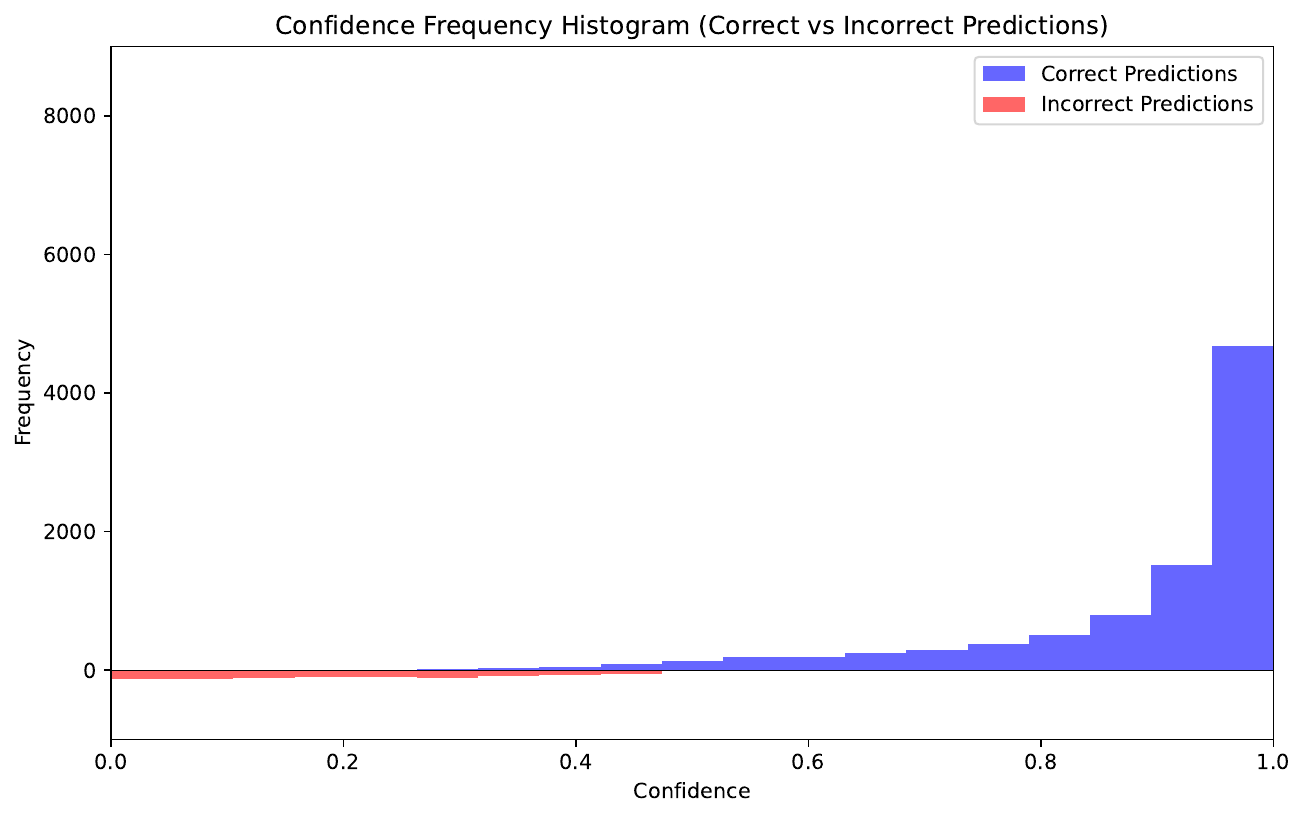}
        \caption{BSCE at epoch 250}
    \end{subfigure}
    \hfill
    \begin{subfigure}[b]{0.43\textwidth}
        \centering
        \includegraphics[width=\textwidth]{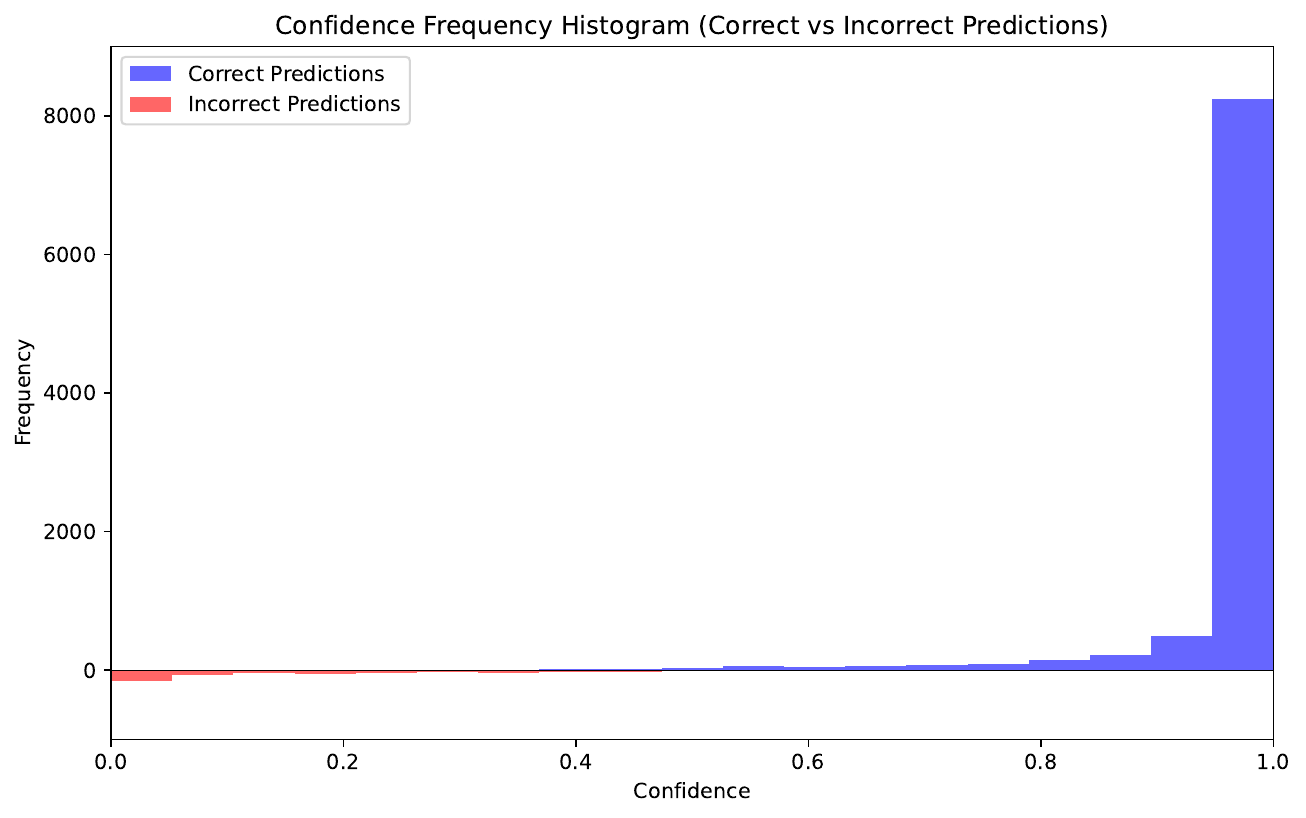}
        \caption{BSCE at epoch 350}
    \end{subfigure}

    \caption{Evolution of confidence distributions under different training epochs for various loss functions. The histograms and density curves illustrate the confidence distributions of different loss functions (FL, DFL, BSCE, BSCE-GRA) during training. This comparison reveals how different loss functions shape the model's confidence throughout the training process.}
    \label{fig: confidence distribution2}
\end{figure*}

We track the number of test samples classified correctly or incorrectly throughout the training process at epochs 50, 150, 250, and 350, as shown in Figure~\ref{fig: confidence distribution1} and Figure~\ref{fig: confidence distribution2}.
The confidence represents the probability assigned to the ground-truth class, and we report the frequency of both correct and incorrect predictions.

These figures provide insight into how different loss functions influence model predictions.
Notably, Cross Entropy tends to produce predictions with high confidence from early on in training.
By the final epoch, Cross Entropy frequently assigns near 100\% confidence to predictions, even when they are incorrect. In contrast, other loss functions impose constraints that limit overconfident predictions.

\begin{table}[]
        \centering
        \begin{tabular}{|c|c|c|c|c|c|c|}
            \hline
            \textbf{Exp} & 3 & 4 & 5 & 6 & 7 & 8 \\ 
            \hline
            \textbf{ECE} & 2.78 & 1.13 & 2.83 & 5.13 & 6.96 & 9.28 \\
            \hline
        \end{tabular}
        \caption{Exponent Comparison on CIFAR100 with ResNet50}\label{table: Exponent}
        \vspace{-5.5mm}
\end{table}

\section{Hyperparameter Selection of BSCE-GRA}
We determine the optimal hyperparameters for BSCE-GRA, including $\gamma$ and $\beta$, using cross-validation, a standard approach as mentioned by~\citep{mukhoti2020calibrating}: "Finding an appropriate $\gamma$ is normally done using cross-validation.
Traditionally, $\gamma$ is fixed for all samples in the dataset."
We observe that the FLSD-53 strategy is employed in~\citep{mukhoti2020calibrating} to better control gradient magnitudes by achieving a more favorable trade-off in the function$u(\hat{p}(\vx))$ in Eq.~\ref{eq: uncertainty-gra loss gradient}, as discussed in the same work. The primary reason we use fixed hyperparameters in our method, rather than an adaptive $\gamma$ strategy, is that BSCE-GRA inherently achieves gradient magnitude control and favorable trade-offs during optimization.
We acknowledge that tuning $\gamma$ can improve calibration performance with Focal Loss. However, our proposed DFL also fulfills the requirements described in~\citep{mukhoti2020calibrating} by incorporating additional logits into the calculation.
Moreover, we provide further empirical results for the selection of $\gamma$ and $\beta$ using ResNet50 on CIFAR-10, including ECE and prediction error, as shown in Figure~\ref{fig: bsce_gamma_ece} and Figure~\ref{fig: bsce_gamma_error}.
We also conduct the exponent hyperparameter comparison on CIFAR100 with ResNet50, the results are provided in Table~\ref{table: Exponent}.

\section{Computation Efficiency}
We further conduct experiments to validate the computation efficiency of BSCE-GRA.
Although BSCE-GRA introduces an additional MSE calculation compared to CE, but does not affect backpropagation and has no significant impact on training time.
We train a ResNet50 on CIFAR10 with default experiment setting.
The running time of CE and BSCE-GRA are 128 and 139 mins, separately.

\begin{table}[]
\centering
        \begin{tabular}{|c|c|c|c|}
            \hline
            \textbf{Loss} & CE & FL & BSCE-GRA \\ 
            \hline
            \textbf{ECE} & 2.07 & 1.16 & 0.93 \\
            \hline
        \end{tabular}
        \caption{ECE performance on ViT among CE, FL and BSCE-GRA}\label{Table: ViT}
\end{table}

\section{Effectiveness on More Model Structure}
To further validate the effectiveness of the proposed method across a wider range of model architectures, we fine-tune a ViT model pretrained on IN-1K for 50 epochs on CIFAR-10 using different loss functions, including CE, FL, and BSCE-GRA. The backbone model is obtained from Hugging Face~\footnote{\hyperlink{https://huggingface.co/google/vit-base-patch16-224.}{https://huggingface.co/google/vit-base-patch16-224.}}, and we follow their fine-tuning guide throughout the process. The results are reported in Table~\ref{Table: ViT}.

\subsection{Sample-wise Calibration Metric}
The proposed framework leverages sample-wise uncertainty as a gradient weight to enhance calibration. However, most existing calibration metrics, such as Expected Calibration Error (ECE), rely on binning strategies, making them unsuitable for directly measuring sample-wise calibration.
A potential solution for evaluating sample-wise calibration is to measure the difference between ground truth and predicted probabilities. Although obtaining accurate ground truth probabilities is challenging, datasets like CIFAR-10H~\cite{peterson2019human} approximate them through human annotations. Predicted probabilities from models may still exhibit bias, but they can be calibrated using methods like "consistency"~\cite{tao2024consistency} or temperature scaling.
\citet{tao2024consistency} perturb the model feature for a sample with a noise several times and consider the expectation of predicted probability as a local consistency of the sample.
If the sample is less certain, the "consistency" will have a high variance.

To further validate the effectiveness of proposed framework, we utilize "1-consistency"~\cite{tao2024consistency} as an uncertainty weight to evaluate the proposed framework. We conduct experiments on CIFAR10 using the ResNet50 architecture, achieving an ECE of 0.98 and an accuracy of 94.7\%.

\begin{table}[]

        \centering
        \begin{tabular}{|c|c|c|c|} 
        \hline 
        \textbf{Loss} & CE & FL & BSCE-GRA \\ 
        \hline 
        \textbf{ECE} & 3.69 (2.26) & 3.28 (2.51) & 2.63 (1.61)\\
        \hline 
        \end{tabular}
        \caption{ECE performance on ImageNet with ResNet50}\label{ImageNet}

\end{table}

\section{Effectiveness on Large Scale Dataset}
To thoroughly evaluate the effectiveness of the proposed method, we fine-tune a ResNet-50 model, pretrained on IN-1K and provided by PyTorch, for 90 epochs on the IN-1K dataset. The fine-tuning process utilizes three different loss functions: Cross Entropy (CE), Focal Loss (FL), and our proposed BSCE-GRA.
For optimization, we employ the Adam optimizer, which effectively balances convergence speed and stability. Additionally, we utilize a Cosine Learning Rate Scheduler to adapt the learning rate throughout training, promoting efficient convergence. The detailed fine-tuning process follows standard practices to ensure fair comparison among different loss functions.
The performance results, including accuracy and calibration metrics, are summarized in Table~\ref{ImageNet}. These results provide insight into the robustness and adaptability of the proposed method across diverse training settings.

\end{document}